\documentclass[10pt,twocolumn,letterpaper]{article}
\usepackage{cvpr}
\usepackage{times}  
\usepackage{helvet}  
\usepackage{courier}  
\usepackage{url}  
\usepackage{graphicx}  

\usepackage[utf8]{inputenc} 
\usepackage[T1]{fontenc}    
\usepackage{booktabs}       
\usepackage{amsfonts}       
\usepackage{nicefrac}       
\usepackage{microtype}      

\usepackage{amsmath}
\usepackage{amssymb}
\usepackage{multirow}
\usepackage{xcolor}
\usepackage{subcaption}
\usepackage[inline]{enumitem}
\setlist{nolistsep}
\usepackage{placeins} 
\usepackage[thinlines]{easytable} 

\cvprfinalcopy 

\def\cvprPaperID{4833} 

\ifcvprfinal\fi

\newcommand{\B}[1]{\textbf{#1}}
\newcommand{\GAN}[1]{\textsc{gan}$_{#1}$}
\newcommand{\Ens}[1]{Ens$_{#1}$}
\newcommand{\Tab}[1]{Table~\ref{#1}}
\newcommand{\Fig}[1]{Figure~\ref{#1}}
\newcommand{\App}[1]{Appendix~\ref{#1}}

\newcommand{\FC}[1]{#1$^\ast$}

\providecommand{\scalT}[2]{\left\langle{#1},{#2}\right\rangle}
\providecommand{\nor}[1]{\left\lVert {#1} \right\rVert}
  
\begin{document}

\title{Adversarial Semantic Alignment for Improved Image Captions}

\author{Pierre Dognin$^*$,  Igor Melnyk$^*$,  Youssef Mroueh$^*$,  Jerret Ross$^*$  \& Tom Sercu$^*$ \\
	IBM Research, Yorktown Heights, NY \\
	{\small\texttt{\{pdognin,mroueh,rossja\}@us.ibm.com}, \texttt{\{igor.melnyk,tom.sercu1\}@ibm.com}}
}

\maketitle

{\let\thefootnote\relax\footnotetext{$^{*}$Equal Contributions. Authors in alphabetical order.}}

\begin{abstract}
  In this paper we study image captioning as a conditional GAN training, proposing both a context-aware LSTM captioner and co-attentive discriminator, which enforces semantic alignment between images and captions.
  We empirically focus on the viability of two training methods: Self-critical Sequence Training (SCST) and Gumbel Straight-Through (ST) and demonstrate that SCST shows more stable gradient behavior and improved results over Gumbel ST, even without accessing discriminator gradients directly.
  We also address the problem of automatic evaluation for captioning models and introduce a new semantic score, and show its correlation to human judgement.
  As an evaluation paradigm, we argue that an important criterion for a captioner is the ability to generalize to compositions of objects that do not usually co-occur together. To this end, we introduce a small captioned Out of Context (OOC) test set.
  The OOC set, combined with our semantic score, are the proposed new diagnosis tools for the captioning community.
  When evaluated on OOC and MS-COCO benchmarks, we show that SCST-based training has a strong performance in both semantic score and human evaluation, promising to be a valuable new approach for efficient discrete GAN training. 
\end{abstract}

\section{Introduction}

Significant progress has been made on the task of generating image descriptions using neural image captioning. Early systems were traditionally trained using cross-entropy (CE) loss minimization \cite{GoogleNIC,Karpathy,CapAttention}. Later, reinforcement learning techniques \cite{Ranzato,scst,spider} based on policy gradient methods, e.g., REINFORCE, were introduced to directly optimize the $n$-gram matching metrics such as CIDEr \cite{CIDEr}, BLEU4 \cite{BLEU} or SPICE \cite{anderson2016spice}. Along a similar idea, \cite{scst} introduced Self-critical Sequence Training (SCST), a light-weight variant of REINFORCE, which produced state of the art image captioning results using CIDEr as an optimization metric.
Although optimizing the above automatic metrics might be a promising direction to take, these metrics unfortunately miss an essential part of the semantic alignment between image and caption.
They do not provide a way to promote naturalness of the language, e.g., as measured by a Turing test, so that the machine-generated text becomes indistinguishable from the text created by humans. 

To address the problem of diversity and naturalness, image captioning has recently been explored in the framework of GANs \cite{goodfellow2014generative}.
The main idea is to train a discriminator to detect a signal on the misalignment between an image and a generated sentence, while the generator (captioner) can use this signal to improve its text generation mechanism to better align the caption with a given image.
Due to the discrete nature of text generation, GAN training remains challenging and has been generally tackled with either reinforcement learning techniques \cite{yu2016seqgan,maligan,HjelmJCCB17,rajeswar2017adversarial, bodai} or by using the Gumbel softmax relaxation \cite{jang2016categorical}, for example, as in \cite{Shetty, Kusner2016GANSFS}. 

Despite these impressive advances, image captioning is far from being a solved task.
It still is a challenge to satisfactory bridge a semantic gap between image and caption, and to produce diverse, creative and human-like captions.
The current captioning systems also suffer from a dataset bias: the models overfit to common objects co-occurring in common context, and they struggle to generalize to scenes where the same objects appear in unseen contexts.
Although the recent advances of applying GANs for image captioning to promote human-like captions is a very promising direction, the discrete nature of the text generation process makes it challenging to train such systems. The results in \cite{bodai, Shetty} are encouraging but the proposed solutions are still complex and computationally expensive. Moreover, the recent work of \cite{caccia18} showed that the task of text generation for the current discrete GAN models is still challenging, many times producing unsatisfactory results, and therefore requires new approaches and methods. Finally, evaluation of image captioning using automated metrics such as CIDEr, BLEU4, etc. is still unsatisfactory since simple $n$-gram matching, that does not reference the image, remains inadequate and sometimes misleading for scoring diverse and descriptive captions.

In this paper, we propose to address the above issues by accomplishing the following three main objectives:  
\begin{enumerate*}[label={\arabic*)},font={\bfseries}]
	\item \textit{Architectural and algorithmic improvements}: We propose a novel GAN-based framework for image captioning that enables better language composition and more accurate compositional alignment of image and text (Section~\ref{sec:cap_disc}), as well as a light-weight and efficient approach for discrete GAN training based on SCST (Section \ref{sec:training}). 
	\item \textit{Automated scoring metric}: We propose the \emph{semantic score}, which enables quantitative automatic evaluation of caption quality and its alignment to the image across multiple models (Section \ref{sec:Evaluation}). 
	\item \textit{Diagnostic dataset}: Finally, we introduce the Out of Context (OOC) test set which is a quick and useful diagnostic tool for checking a model's generalization to out of context scenes (Section \ref{sec:Evaluation}).
\end{enumerate*}

\section{Adversarial Caption Generation}

In this Section we present our novel captioner and the discriminator models. We employ SCST for discrete GAN optimization and compare it with the approach based on the Gumbel trick.
Our experiments (Section \ref{sec:expres}) show that SCST obtains better results, even though it does not directly access the discriminator gradients.

\subsection{Compositional Captioner and Discriminator} \label{sec:cap_disc}

Here we introduce an image captioning model with attention that we call \emph{context aware captioning} based on \cite{sentinel}. 
This allows the captioner to compose sentences based on fragments of observed visual scenes in the training.
Furthermore, we introduce a discriminator that scores the alignment between images and captions based on a co-attention model \cite{coattention}.
This gives the generator a signal on the semantic alignment and the compositional nature of visual scenes and language.
We show in Section \ref{sec:expres} that we obtain better results across evaluation metrics when using this co-attentive discriminator.

\begin{figure}[ht!]
	\centering
	\includegraphics[width=0.6\linewidth]{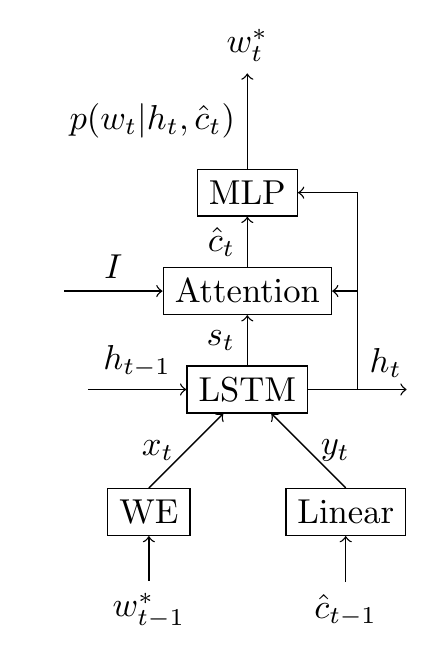} 
	\caption{Context Aware Captioner. At each step $t$, the textual information $w^*_{t-1}$, and the mixture of image features and visual sentinel $\hat{c}_{t-1}$ from previous step $t\!-\!1$ are fed to the LSTM to make it aware of past attentional contexts.}
	\label{fig:context_aware}
\end{figure}

\noindent \textbf{Context Aware Captioner $\boldsymbol{G_{\theta}}$}. 
For caption generation, we use an LSTM with visual attention \cite{CapAttention,scst} together with a visual sentinel \cite{sentinel} to give the LSTM a choice of attending to visual or textual cues.
While \cite{sentinel} feeds only an average image feature to the LSTM at each step, we feed a mixture of image and visual sentinel features $\hat{c}_{t-1}$ from the previous step to make the LSTM aware of the last attention context, as seen in \Fig{fig:context_aware}.
We call it \emph{Context Aware} Attention.
This simple modification gives significant gains, as the captioner is now aware of the contextual information used in the past.
As reported in Table~\ref{tab:attperff}, a captioner with an adaptive visual sentinel \cite{sentinel} gives 99.7 CIDEr vs. 103.3 for our Context Aware Attention on COCO validation set.

\begin{table}[!htb]
	\centering
	\begin{tabular}{lrr}
		\toprule
		Attention Model & CE    & RL    \\
		\cmidrule(r){1-1}
		\cmidrule(r){2-3}
		Att2All \cite{scst}     & 98.5  & 115.7 \\
		Sentinel \cite{sentinel} & 99.7  &       \\
		Context Aware (ours)   & 103.3 & 118.6 \\
		\bottomrule
	\end{tabular}
	\caption{CIDEr performance of captioning systems given various attention mechanisms, Att2All \cite{scst}, sentinel attention \cite{bodai} and
		Context Aware attention on COCO validation set. Models are built using cross-entropy (CE) and SCST \cite{scst} (RL). Context aware attention brings large gains in CIDEr for both CE and RL trained models.}
	\label{tab:attperff}
\end{table}

\begin{figure*}[ht!]
	\centering	
	\includegraphics[width=0.85\linewidth]{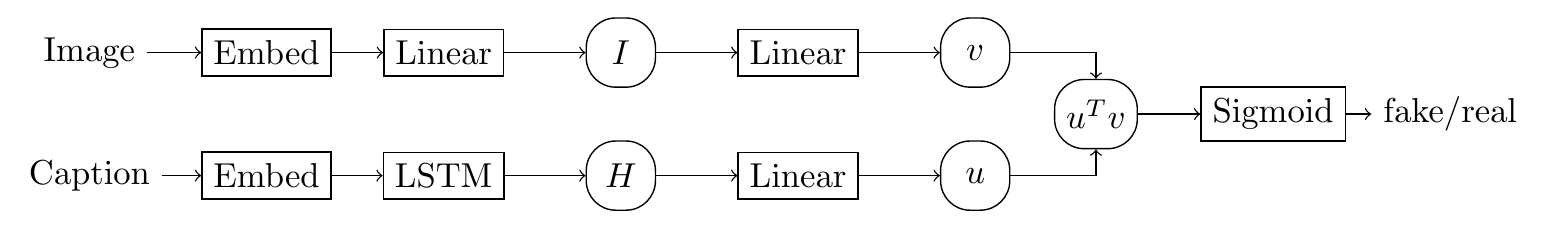} \\
	(a)  Joint-Embedding Discriminator (Joint-Emb)  \cite{bodai} \\
	\includegraphics[width=\linewidth]{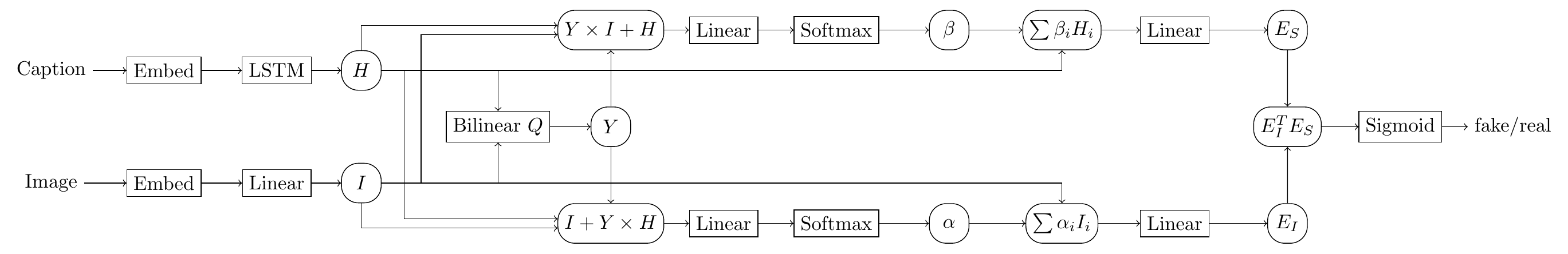} \\
	(b)  Proposed Co-Attention Discriminator (Co-att)
	\caption{Discriminator architectures.
		\textbf{(a)} Joint-Embedding Discriminator from \cite{bodai}.
		\textbf{(b)} Our proposed $D_{\eta}$. By jointly embedding the image and caption with a co-attention model, we give the discriminator the ability to modulate the image features depending on the caption and vice versa.  } \label{fig:discArch}
\end{figure*}

\noindent \textbf{Co-attention Pooling Discriminator $\boldsymbol{D_{\eta}}$}.
The task of the discriminator is to score the similarity between an image and a caption. Previous works jointly embed the modalities at the similarity computation level, which we call \emph{late joint embedding}, see \Fig{fig:discArch}~(a). Instead, we propose to jointly embed image and caption in earlier stages using a co-attention model \cite{coattention,Cicero} and compute similarity on the attentive pooled representation. We call this a \emph{Co-attention} discriminator (\Fig{fig:discArch}~(b)) and provide architectural details below.

\begin{figure*}[ht!]
	\centering
	\includegraphics[width=0.7\linewidth]{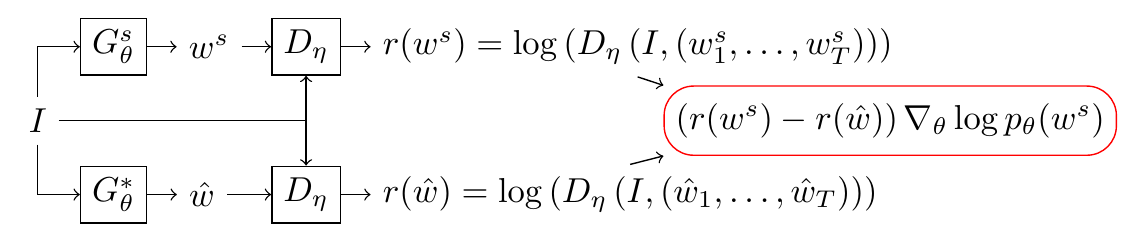}
	\caption{SCST Training of GAN-captioning. }
	\label{fig:scstgan}
\end{figure*}

Given a sentence $w$ composed of a sequence of words $(w_1,\dotsc w_T)$, the discriminator embeds each word using the LSTM (state dimension $m\!=\!512$) to get $H=[h_1,\dotsc h_T]^\top$ for $H \in \mathbb{R}^{T\times m}$, where $h_t,c_t= \mathrm{LSTM}(h_{t-1},c_{t-1},w_{t})$. For image $I$, we extract features $(I_1,\dotsc I_C)$, where $C\!=\!14\!\times\!14\!=\!196$ (number of crops) and also embed them as $I=[WI_1,\dots WI_{C}]^\top \in \mathbb{R}^{C\times m}$,
where $W\in \mathbb{R}^{m\times d_{I}}$, and $d_{I}=2048$, our image feature size.
Following \cite{coattention}, we then compute a correlation $Y$ between image and text using bilinear projection
$Q \in \mathbb{R}^{m\times m}$, $Y=\tanh(IQH^\top) \in \mathbb{R}^{C\times T}$. Matrix $Y$ is used to compute co-attention weights of one modality conditioned on another:
\begin{align*}
  \alpha &=\mathrm{Softmax}(\mathrm{Linear}(\tanh(I W_I +Y H W_{Ih} ))) \in \mathbb{R}^{C}, \\
  \beta&=\mathrm{Softmax}(\mathrm{Linear}(\tanh(H W_h + Y^\top I W_{hI} ))) \in \mathbb{R}^{T},
\end{align*}
where all new matrices are in $\mathbb{R}^{m\times m}$. The above weights are used then to combine the word and image features: $E_{I} = U_I \left(\sum_{i=1}^{C} \alpha_i WI_i \right)$ and $ E_{S} = V_S \left(\sum_{j=1}^T \beta_j h_j \right)$
for $U_I,V_S \in \mathbb{R}^{m\times m}$.
Finally, the image-caption score is computed as $D_{\eta}(I,w)=\rm{Sigmoid}(E^{\top}_{I}E_{S})$ ($\eta$, discriminator parameters).
In Section \ref{sec:expres} we compare  $D_{\eta}$ with the late joint embedding approach of \cite{bodai,Shetty}, where $E_{I}$ is the average spatial pooling of CNN features and $E_{S}$
the last state of LSTM. We refer to this discriminator as \emph{Joint-Emb} and to ours as \emph{Co-att} (see \Fig{fig:discArch}).

\subsection{Adversarial Training}\label{sec:training}

In this Section we describe the details of the adversarial training of the discriminator and the captioner.

\noindent \textbf{Training $D_{\eta}$}. 
Our discriminator $D_{\eta}$ is not only trained to distinguish real captions from fake (generated), but also to detect when images are coupled with random unrelated real sentences, thus forcing it to check not only the sentence composition but also the semantic relationship between image and caption.
To accomplish this, we solve the following optimization problem:  $\max_{\eta} \mathcal{L}_{D}(\eta)$, where the loss $\mathcal{L}_{D}(\eta)$
\begin{align}
  \mathbb{E}_{I,{w \in S(I)}} \log &D_{\eta}(I,w) + \frac{1}{2}\mathbb{E}_{I,w^s \sim p_{\theta}(.|I)} \log\left(1\!-\!D_{\eta}(I, w^s)\right) \nonumber \\
                                        &+ \frac{1}{2}\mathbb{E}_{I,{w'}\notin S(I)} \log\left( 1- D_{\eta}(I,w')\right) \label{eq:LossD},
\end{align}
where $w$ is the real sentence,  $w^s$ is sampled from generator $G_{\theta}$ (fake caption), and $w'$ is a real but randomly picked caption.

\noindent \textbf{Training $\boldsymbol{G_{\theta}}$}.
The generator is optimized to solve $\max_{\theta}\mathcal{L}_{G}(\theta)$, where $\mathcal{L}_{G}(\theta) \! = \! \mathbb{E}_{I}\log  D_{\eta}(I,G_{\theta}(I))$.
The main difficulty is the discrete, non-differentiables nature of the problem.
We propose to solve this issue by adopting SCST \cite{scst}, a light-weight variant of the policy gradient method,
and compare it to the Gumbel relaxation approach of \cite{jang2016categorical}.

\noindent \textbf{Training $\boldsymbol{G_{\theta}}$ with SCST}. SCST \cite{scst} is a REINFORCE variant that uses the reward under the decoding algorithm as baseline.
In this work, the decoding algorithm is a ``greedy max'', selecting at each step the most probable word from $\arg\max p_{\theta}(.|h_t)$.
For a given image, a single sample $w^s$ of the generator is used to estimate the full sequence reward, $\mathcal{L}^{I}_{G}(\theta)= \log(D(I,w^s))$ where $w^s\sim p_{\theta}(.|I)$.
Using SCST, the gradient is estimated as follows:
\begin{align*}
  \nabla_{\theta}\mathcal{L}^I_{G}(\theta) \! &\approx \! (\log  D_{\eta}(I,w^s) - \underbrace{\log  D_{\eta}(I,\hat{w})}_{\text{Baseline}}) \nabla_{\theta}\log p_{\theta}(w^s|I) \\
  &=  \left(\log \frac{ D_{\eta}(I,w^s)}{D_{\eta}(I,\hat{w})}\right) \nabla_{\theta}\log p_{\theta}(w^s|I),
\end{align*}
where $\hat{w}$ is obtained using \emph{greedy max} (see \Fig{fig:scstgan}). Note that the baseline does not change the expectation of the gradient but reduces the variance of the estimate. 

Also, observe that the GAN training can be regularized with any NLP metric $r_{\mathrm{\scriptscriptstyle NLP}}$ (such as CIDEr) to enforce closeness of the generated captions to the provided ground truth on the $n$-gram level; the gradient then becomes:
\begin{align*}
\left(\!\log \! \frac{ D_{\eta}(I,w^s)}{D_{\eta}(I,\hat{w})}\! + \! \lambda \left( r_{\mathrm{\scriptscriptstyle NLP}}(w^s)\!-\!r_{\mathrm{\scriptscriptstyle NLP}}(\hat{w}) \right) \! \right) \! \nabla_{\theta} \! \log p_{\theta}(w^s|I).
\end{align*}

There are two main advantages of SCST over other policy gradient methods used in the sequential GAN context:
\textbf{1)} The reward in SCST can be global at the sentence level and the training still succeeds. In other policy gradient methods, e.g., \cite{bodai,spider},
the reward needs to be defined at each word generation with the \emph{full} sentence  sampling, so that the discriminator needs to be evaluated $T$ times (sentence length).
\textbf{2)} In \cite{bodai,spider,HjelmJCCB17}, many Monte-Carlo rollouts are needed to reduce variance of gradients, requiring many forward-passes through the generator.
In contrast, due to a strong baseline, only a single sample estimate is enough in SCST.

\paragraph{Training $G_{\theta}$ : Gumbel Trick.} An alternative way to deal with the discreteness of the generator is by using Gumbel re-parameterization \cite{jang2016categorical}.
Define the soft samples $y_t^j$, for $t=1,\dots T$ (sentence length) and $j=1,\dotsc K$ (vocabulary size)  such that:
$ y^j_t=\mathrm{Softmax}\left(\frac{1}{\tau}(\mathrm{logits}_{\theta}(j |h_t,I)+g_j )\right),$
where $g_j$ are samples from Gumbel distribution, $\tau$ is a temperature parameter.
We experiment with the Gumbel Soft and Gumbel Straight-Through (Gumbel ST) approaches, recently used in \cite{Shetty,Kusner2016GANSFS}.

For \emph{Gumbel Soft}, we use the soft samples $y_t$ as LSTM input $w_{t+1}^s$ at the next time step and in $D_{\eta}$:
\[
\nabla_{\theta}\mathcal{L}^I_{G}(\theta)= \nabla_{\theta} \log(D_{\eta}(I,(y_1,\dots y_T))).
\]
For \emph{Gumbel ST}, we define one-hot encodings $\mathcal{O}_t\!=\!\mathrm{OneHot}(\arg\max_j y_t^j)$ and approximate the gradients $\partial \mathcal{O}_t^j / \partial y_t^{j'} = \delta_{j j'}$.
To sample from $G_{\theta}$ we use the hard $\mathcal{O}_t$  as LSTM input $w_{t+1}^s$ at the next time step and in $D_{\eta}$, hence the gradient becomes:
\[
\nabla_{\theta}\mathcal{L}^I_{G}(\theta)= \nabla_{\theta} \log(D_{\eta}(I,(\mathcal{O}_1,\dots \mathcal{O}_T))).
\]

\noindent Observe that this loss can be additionally regularized with Feature Matching (FM) as follows:
\begin{align}
  &\mathcal{L}^I_{G}(\theta) =  \log(D_{\eta}(I,(y_1,\dots y_T))) \nonumber \\
  &-\lambda^I_{F}\left(||E_{I}(w^*_1,\dots w^*_T)-E_{I}(y_1,\dots y_T) ||^2 \right) \nonumber \\
   &-\lambda^S_{F}\left(||E_{S=(w^*_1,\dots w^*_T)}(I)-E_{S=(y_1,\dots y_T)}(I) ||^2 \right),
    \label{eq:FM}
\end{align}
where $(w^*_1,\dots w^*_T)$ is the ground truth caption corresponding to image $I$, and $E_{I}$ and $E_S$ are co-attention image and sentence embeddings (as defined in Section \ref{sec:cap_disc}).
Feature matching enables us to incorporate more granular information from discriminator representations of the ground truth caption,
similar to how SCST reward can be regularized with CIDEr, computed with a set of baseline captions.


\section{Evaluation: Semantic Score and OOC Set}\label{sec:Evaluation}

\begin{table*}[!htb]
	\centering
	\resizebox{\linewidth}{!}{
		\begin{tabular}{lrrrrrrrrrrrrrrrr}
			\toprule
			{} & \multicolumn{8}{l}{COCO Test Set} & \multicolumn{8}{l}{OOC (Out of Context)} \\
			\cmidrule(r){2-9} \cmidrule(r){10-17}
			{} & \multicolumn{2}{l}{CIDEr} & \multicolumn{2}{l}{METEOR} & \multicolumn{2}{l}{Semantic} & \multicolumn{2}{l}{Vocabulary} & \multicolumn{2}{l}{CIDEr} & \multicolumn{2}{l}{METEOR} & \multicolumn{2}{l}{Semantic} & \multicolumn{2}{l}{Vocabulary} \\
			{} & \multicolumn{2}{l}{}      & \multicolumn{2}{l}{}       & \multicolumn{2}{l}{Score}    & \multicolumn{2}{l}{Coverage}   & \multicolumn{2}{l}{}      & \multicolumn{2}{l}{}       & \multicolumn{2}{l}{Score}    & \multicolumn{2}{l}{Coverage}   \\
			\cmidrule(r){1-1} \cmidrule(r){2-9} \cmidrule(r){10-17}
			CE                                                        & 101.6     & $\pm$0.4 & 0.260     & $\pm$.001 & 0.186     & $\pm$.001 & 9.2      & $\pm$0.1 & 42.2     & $\pm$0.6 & 0.169     & $\pm$.001 & 0.118     & $\pm$.001 & 2.8     & $\pm$0.1 \\
			CIDEr-RL                                                  & \B{116.1} & $\pm$0.2 & 0.269     & $\pm$.000 & 0.184     & $\pm$.001 & 5.1      & $\pm$0.1 & 45.0     & $\pm$0.6 & 0.170     & $\pm$.003 & 0.117     & $\pm$.002 & 2.1     & $\pm$0.0 \\
			\cmidrule(r){1-1} \cmidrule(r){2-9} \cmidrule(r){10-17}                                                                                                                                                                                             
			GAN$_{1}$(SCST, Co-att, $\log(D)$)                        & 97.5      & $\pm$0.8 & 0.256     & $\pm$.001 & 0.190     & $\pm$.000 & 11.0     & $\pm$0.1 & 41.0     & $\pm$1.6 & 0.168     & $\pm$.003 & \B{0.124} & $\pm$.000 & 3.2     & $\pm$0.1 \\
			GAN$_{2}$(SCST, Co-att, $\log(D)\!+\!5\times$CIDEr)       & 111.1     & $\pm$0.7 & \B{0.271} & $\pm$.002 & \B{0.192} & $\pm$.000 & 7.3      & $\pm$0.2 & \B{45.8} & $\pm$0.9 & \B{0.173} & $\pm$.001 & 0.122     & $\pm$.002 & 2.8     & $\pm$0.1 \\
			GAN$_{3}$(SCST, Joint-Emb, $\log(D)$)                     & 97.1      & $\pm$1.2 & 0.256     & $\pm$.002 & 0.188     & $\pm$.000 & 11.2     & $\pm$0.1 & 41.8     & $\pm$1.6 & 0.167     & $\pm$.002 & 0.122     & $\pm$.001 & 3.3     & $\pm$0.0 \\
			GAN$_{4}$(SCST, Joint-Emb, $\log(D)\!+\!5\times$CIDEr)    & 108.2     & $\pm$4.9 & 0.267     & $\pm$.004 & 0.190     & $\pm$.000 & 8.3      & $\pm$1.6 & 45.4     & $\pm$1.4 & 0.173 & $\pm$.002 & 0.122     & $\pm$.003 & 2.8     & $\pm$0.2 \\
			\cmidrule(r){1-1} \cmidrule(r){2-9} \cmidrule(r){10-17}                                                                                                                                                                                              
			GAN$_{5}$(Gumbel Soft, Co-att, $\log(D)$)                 & 93.6      & $\pm$3.3 & 0.253     & $\pm$.007 & 0.187     & $\pm$.002 & 11.1     & $\pm$1.2 & 38.3     & $\pm$3.7 & 0.164     & $\pm$.006 & 0.121     & $\pm$.004 & 3.3     & $\pm$0.3 \\
			GAN$_{6}$(Gumbel ST, Co-att, $\log(D)$)                   & 95.4      & $\pm$1.5 & 0.249     & $\pm$.004 & 0.184     & $\pm$.003 & 10.1     & $\pm$0.9 & 38.5     & $\pm$1.9 & 0.161     & $\pm$.005 & 0.116     & $\pm$.004 & 3.0     & $\pm$0.2 \\
			GAN$_{7}$(Gumbel ST, Co-att, $\log(D)+$FM)                & 92.1      & $\pm$5.4 & 0.243     & $\pm$.011 & 0.175     & $\pm$.006 & 8.6      & $\pm$0.8 & 36.8     & $\pm$2.3 & 0.157     & $\pm$.006 & 0.110     & $\pm$.005 & 2.5     & $\pm$0.2 \\
			\cmidrule(r){1-1} \cmidrule(r){2-9} \cmidrule(r){10-17}
			\FC{CE}  -- \FC{}denotes \emph{non-attentional} models  & 87.6      & $\pm$1.2 & 0.242     & $\pm$.001 & 0.175     & $\pm$.002 & 9.9      & $\pm$0.8 & 32.0     & $\pm$0.4 & 0.152     & $\pm$.002 & 0.103     & $\pm$.002 & 2.6     & $\pm$.1  \\
			\FC{CIDEr-RL}                                             & 100.4     & $\pm$7.9 & 0.253     & $\pm$.006 & 0.173     & $\pm$.002 & 6.8      & $\pm$1.4 & 33.4     & $\pm$1.4 & 0.154     & $\pm$.003 & 0.101     & $\pm$.003 & 2.1     & $\pm$.2  \\
			\cmidrule(r){1-1} \cmidrule(r){2-9} \cmidrule(r){10-17}                                                                                                                           
			\FC{GAN$_{1}$}(SCST, Co-att, $\log(D)$)                   & 89.7      & $\pm$0.9 & 0.246     & $\pm$.001 & 0.184     & $\pm$.001 & \B{13.2} & $\pm$0.2 & 30.8     & $\pm$1.0 & 0.155     & $\pm$.003 & 0.111     & $\pm$.001 & 3.4     & $\pm$0.1 \\
			\FC{GAN$_{2}$}(SCST, Co-att, $\log(D)+5\times$CIDEr)      & 103.1     & $\pm$0.5 & 0.261     & $\pm$.001 & 0.183     & $\pm$.001 & 7.1      & $\pm$0.2 & 33.7     & $\pm$1.9 & 0.157     & $\pm$.001 & 0.108     & $\pm$.001 & 2.7     & $\pm$0.1 \\
			\FC{GAN$_{3}$}(SCST, Joint-Emb, $\log(D)$)                & 90.7      & $\pm$0.1 & 0.248     & $\pm$.001 & 0.181     & $\pm$.001 & 12.9     & $\pm$0.1 & 30.8     & $\pm$2.1 & 0.153     & $\pm$.002 & 0.108     & $\pm$.001 & \B{3.5} & $\pm$0.1 \\
			\FC{GAN$_{4}$}(SCST, Joint-Emb, $\log(D)+5\times$CIDEr)   & 102.7     & $\pm$0.4 & 0.260     & $\pm$.001 & 0.182     & $\pm$.001 & 7.7      & $\pm$0.1 & 33.3     & $\pm$2.4 & 0.157     & $\pm$.004 & 0.106     & $\pm$.000 & 2.7     & $\pm$0.1 \\
			\cmidrule(r){1-1} \cmidrule(r){2-9} \cmidrule(r){10-17}
			G-GAN \cite{bodai} from Table 1                           & 79.5      & --       & 0.224     & --        & --        & --        & --       & --       & --       & --       & --        & --        & --        & --        & --      & --       \\
			\bottomrule
		\end{tabular}    
	} 
	\caption{Results for all models mentioned in this work. Scores are reported for both COCO and OOC sets. All results are averaged ($\pm$~standard deviation) over 4 models trained with different random seeds. See \Tab{tab:GANFULL} in \App{app:fulltables} for a full set of results.}
	\label{tab:GAN}
\end{table*}

\textbf{Semantic Score.}
Traditional automatic language metrics, such as CIDEr or BLEU4, are inadequate for evaluating GAN-based image caption models.
As an early alternative, \cite{bodai,Shetty} used GAN discriminator for evaluation, but this is not a fair comparison across models since the GAN generator was trained to maximize the discriminator likelihood.
In order to enable automatic evaluation across models we propose the \emph{semantic score}.
Analogous to ``Inception Score'' \cite{salimans2016improved} for image generation, leveraging a large pretrained classification network,
the semantic score relies on a powerful model, trained with supervision, to heuristically evaluate caption quality and its alignment to the image.
In Section~\ref{sec:expres} we show that our semantic score correlates well with human judgement across metrics, algorithms and test sets.

The semantic score is based on a Canonical Correlation Analysis (CCA) retrieval model \cite{assymetric} which brings the image into the scoring loop by training on the combination of COCO \cite{MSCOCO} and SBU \cite{SBU} ($\sim$1M images), ensuring a larger exposure of the score to diverse visual scenes and captions, and lowering the COCO dataset bias. The semantic score is a cosine similarity in CCA space based on a 15k dimension image embedding from resnet-101 \cite{he15deepresidual}, and a sentence embedding computed using a Hierarchical Kernel Sentence Embedding \cite{assymetric} based on word2vec \cite{word2vec}:

\[
s(x,y) = \frac{\scalT{ \Sigma U^{\top}x}{  V^{\top} y}}{\nor{ \Sigma U^{\top}x}_2\nor{V^{\top} y^*}_2},
\]
where $x$ and $y$ are caption and image embedding vectors, respectively; $U$, $\Sigma$, and $V$ are matrices obtained from CCA as described in details in \cite{assymetric}. Note that the use of word2vec allows the computation of scores for captions whose words fall outside of the COCO vocabulary. The computed score can be interpreted as a likelihood of the image given a caption, it also penalizes the sentences which mention non-existent attributes or objects. See \Tab{tab:sem_sentence_table} in Appendix~\ref{app:SemanticScore} for examples. 

\textbf{Out of Context Set (OOC).} An important property of the captioner is the ability to generalize to images with objects falling outside of their common contexts. 
In order to test the compositional and generalization properties to out-of-context scenes (see \Fig{fig:goodcap_joint} for an example), we expanded the original set of \cite{Jinchoi_contextmodels} (containing $218$ images) to a total of $269$ images and collected $5$ captions per image on Amazon MTurk.  We call the resulting dataset the Out of Context (OOC) set. We note that although the size of OOC set is not large, its main purpose is to be a useful quick diagnostic tool rather than a traditional dataset. The evaluation on OOC is a good indicator of a captioner’s generalization: poor performance is a sign that the model is over-fitted to the training context. Improving OOC scores remains an open area for future work, and we plan to release the OOC set as well as the scripts for computing the semantic score.

\section{Experiments }\label{sec:expres}

\noindent \textbf{Experimental Setup.} We evaluate our proposed method and the baselines on COCO dataset \cite{MSCOCO} (vocabulary size is 10096) using data splits from \cite{Karpathy}:
training set of 113k images with $5$ captions each, validation and test sets 5k each; as well as on the proposed OOC diagnostic set.
Each image is encoded by a resnet-101 \cite{he15deepresidual} without rescaling or cropping, followed
by a spatial adaptive max-pooling to ensure a fixed size of 14$\!\times\!$14$\!\times\!$2048.
An attention mask is produced over the 14$\!\times\!$14 spatial locations, resulting in a spatially averaged 2048-dimension representation.
LSTM hidden state, image, word, and attention embedding dimensions are fixed to 512 for all models. Before the GAN training, all the models are first pretrained with cross entropy (CE) loss.
We report standard language evaluation metrics, the proposed semantic score, and the vocabulary coverage (percentage of vocabulary used at generation).

\begin{figure*}[!ht]
	\centering
	\begin{subfigure}[t]{.4\linewidth}
		\centering
		\includegraphics[width=.8\linewidth]{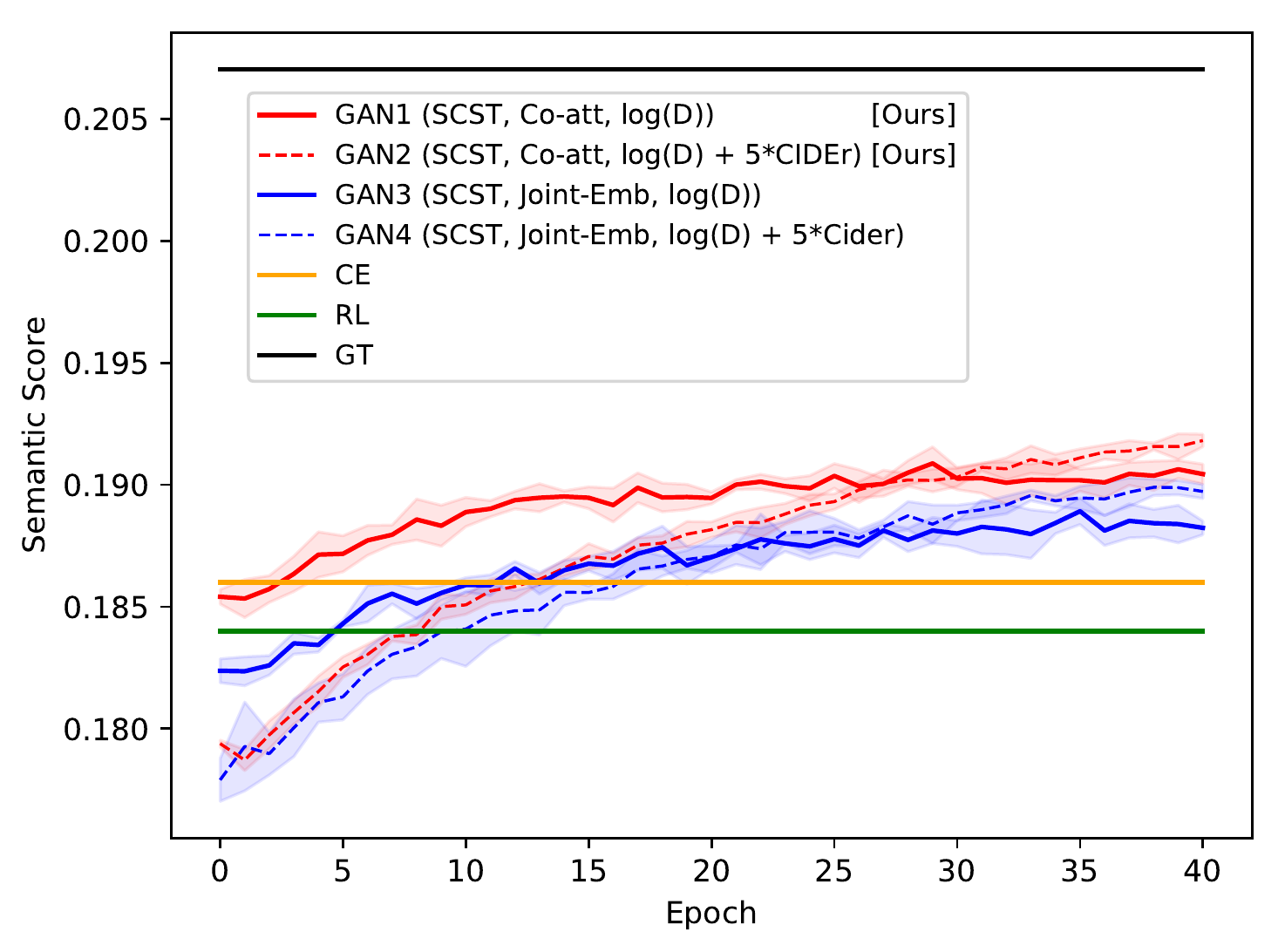}
		\caption{COCO Test}
		\label{fig:sem_test}
	\end{subfigure}%
	\begin{subfigure}[t]{.4\linewidth}
		\centering
		\includegraphics[width=.8\linewidth]{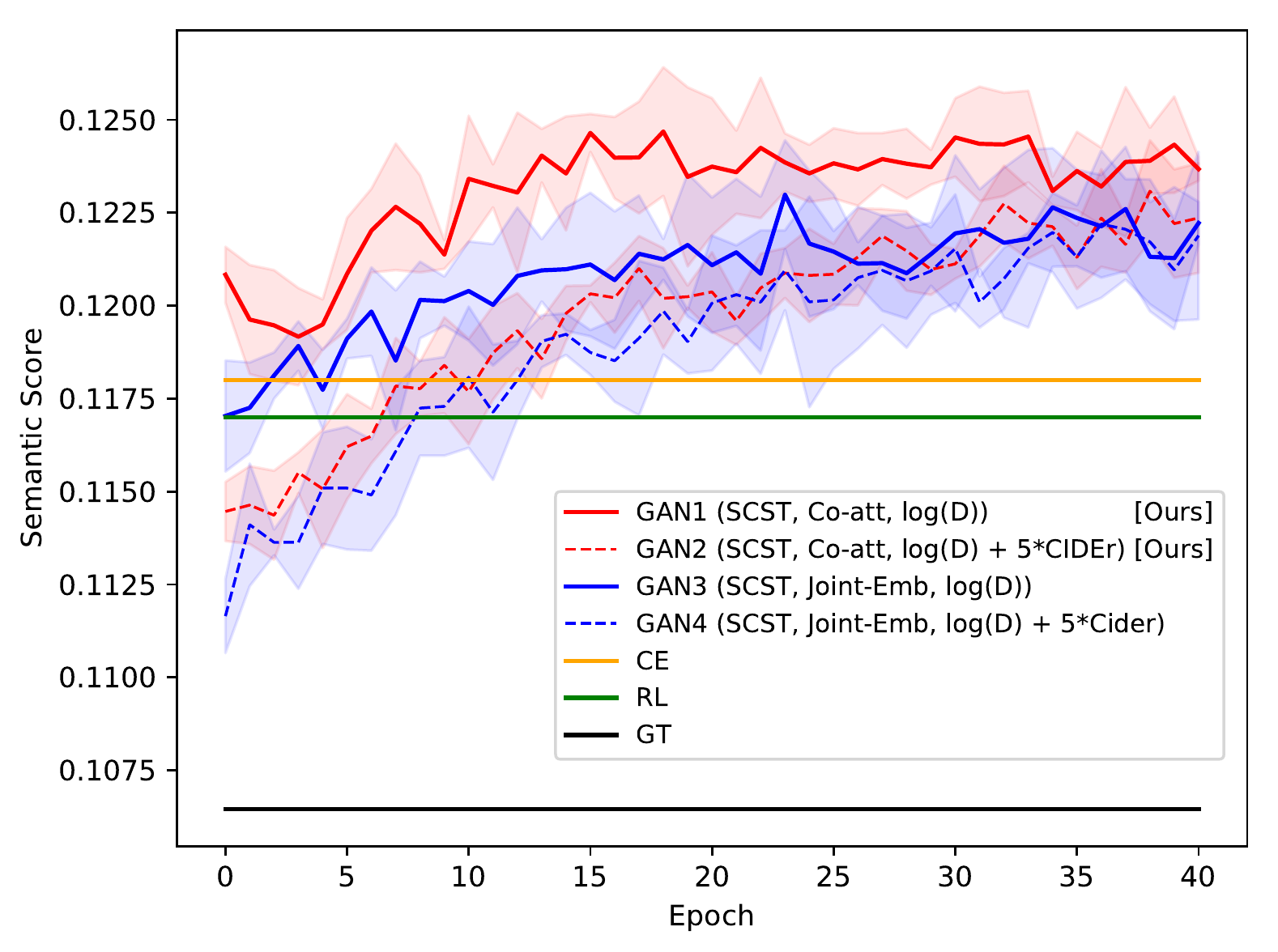}
		\caption{OOC}
		\label{fig:sem_ooc}
	\end{subfigure}%
	\caption{Evolution of semantic scores over training epochs for COCO Test and OOC datasets. Our Co-att models achieve consistently higher scores than CE, RL and Joint-Emb models \cite{bodai}.}
	\label{fig:sem_epoch_joint}
\end{figure*}

\begin{figure*}[!ht]
	\centering
	\begin{subfigure}[t]{.4\linewidth}
		\centering
		\includegraphics[width=.8\linewidth]{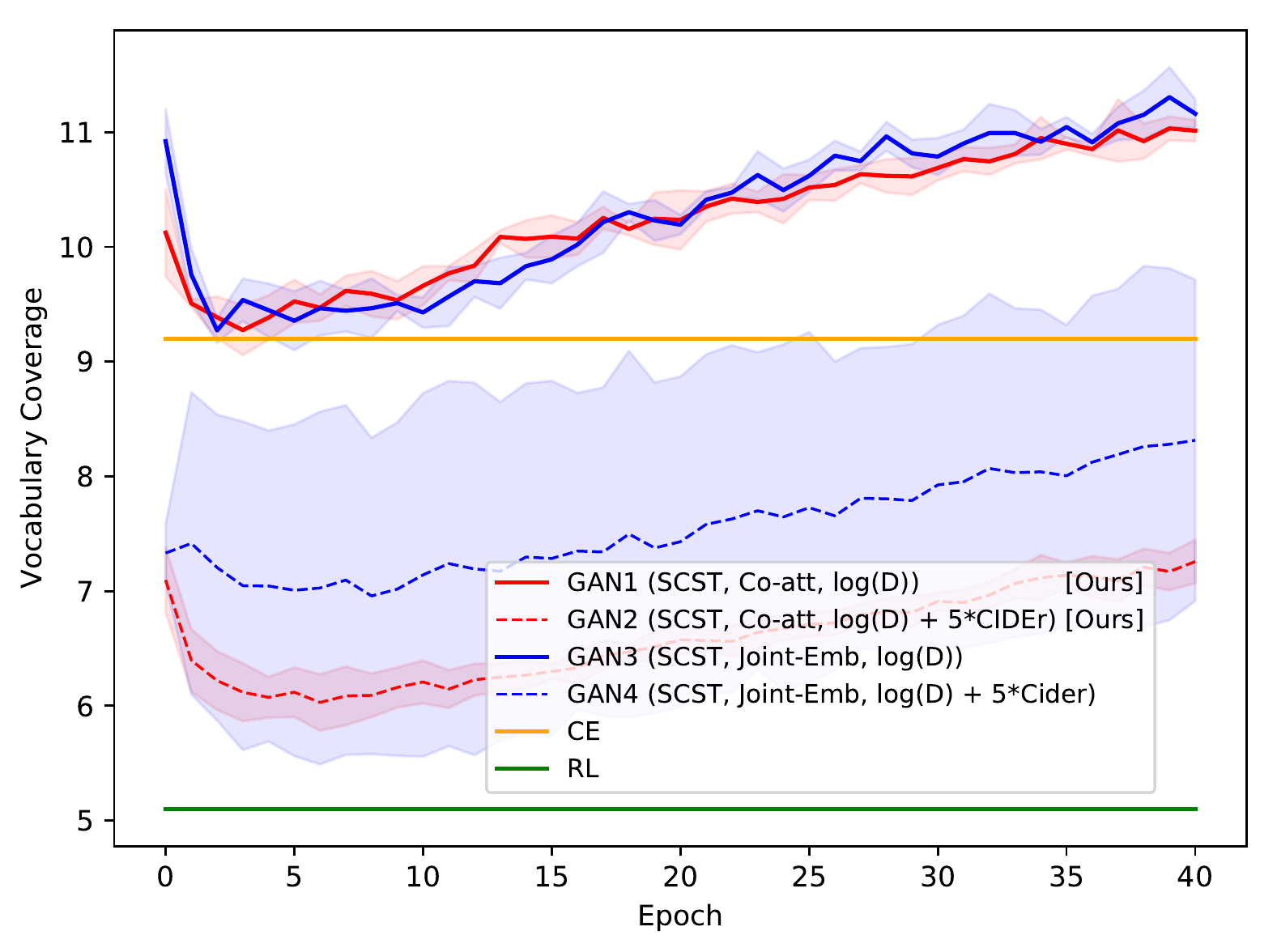}
		\caption{COCO Test}
		\label{fig:voc_test}
	\end{subfigure}%
	\begin{subfigure}[t]{.4\linewidth}
		\centering
		\includegraphics[width=.8\linewidth]{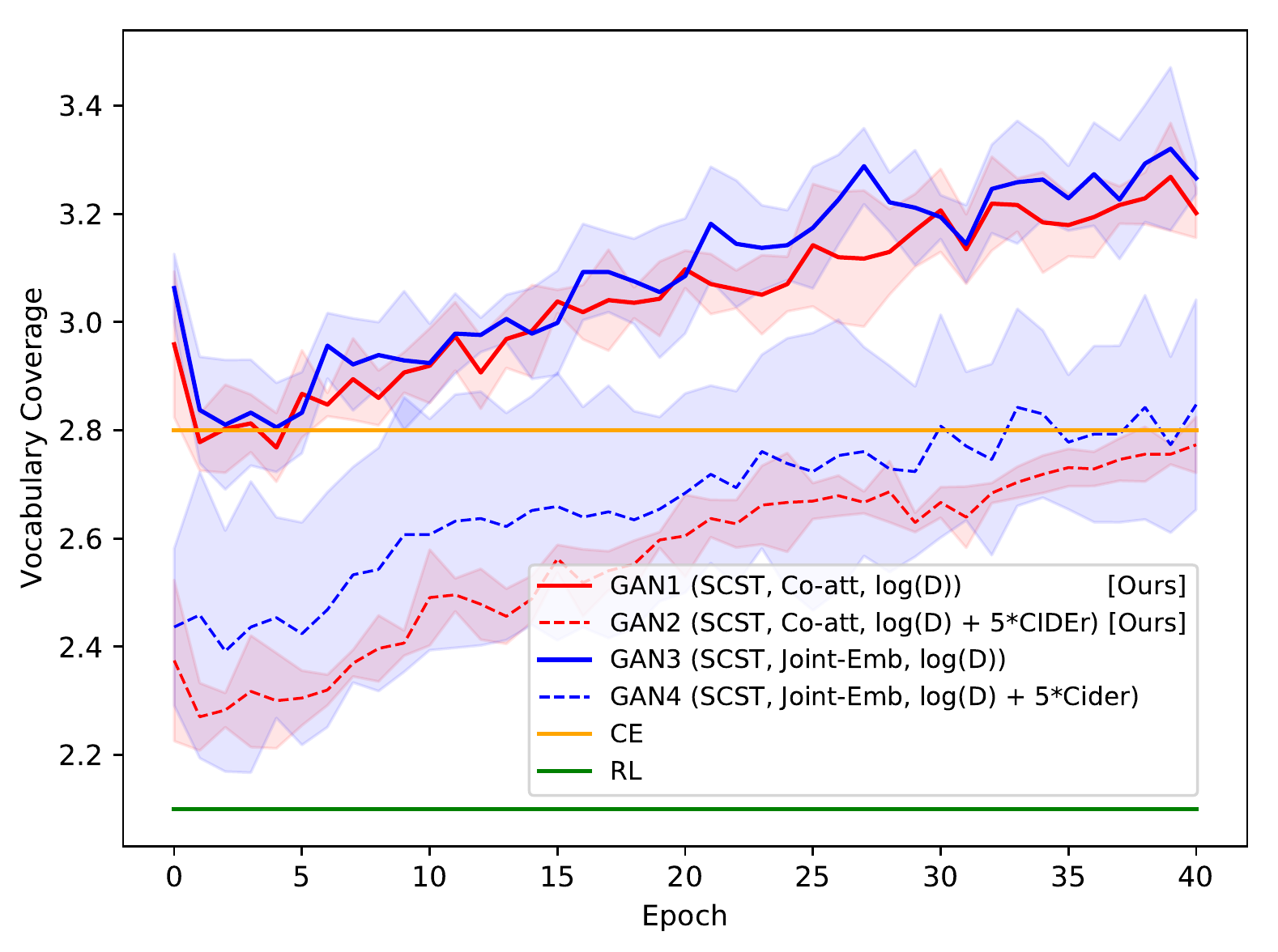}
		\caption{OOC}
		\label{fig:voc_ooc}
	\end{subfigure}%
	\caption{Evolution of vocabulary coverage over training epochs for COCO and OOC datasets.
		As training progresses, we see a correlation between vocabulary coverage and semantic scores for all models.
		Models without CIDEr-regularized SCST GAN rewards achieve best vocabulary coverage.}
	\label{fig:sem_voc_joint}
\end{figure*}

\begin{table*}[!htb]
	\centering
	\resizebox{\textwidth}{!}{
		\begin{tabular}{llrrrrrrrr}
			\toprule
			{} & {} & \multicolumn{4}{l}{COCO Test Set} & \multicolumn{4}{l}{OOC (Out of Context)}  \\ 
			\cmidrule(r){3-6} \cmidrule(r){7-10}
			{} & {} & \multicolumn{1}{l}{CIDEr} & \multicolumn{1}{l}{METEOR} & \multicolumn{1}{l}{Semantic} & \multicolumn{1}{l}{Vocabulary} & \multicolumn{1}{l}{CIDEr} & \multicolumn{1}{l}{METEOR} & \multicolumn{1}{l}{Semantic} & \multicolumn{1}{l}{Vocabulary} \\
			{} & {} & {}                        &                            & \multicolumn{1}{l}{Score}    & \multicolumn{1}{l}{Coverage}   & \multicolumn{1}{l}{}      & \multicolumn{1}{l}{}       & \multicolumn{1}{l}{Score}    & \multicolumn{1}{l}{Coverage}   \\ 
			\cmidrule(r){1-2} \cmidrule(r){3-6} \cmidrule(r){7-10}
			\multirow{ 2}{*}{(CE and RL Baselines)}     & Ens$_{\text{CE}}$(CE)                      & 105.8     & 0.266     & 0.189      & 8.4       & 44.8       & 0.172       & 0.122       & 2.6     \\ 
			& Ens$_{\text{RL}}$(CIDEr-RL)                & \B{118.9} & 0.273     & 0.186      & 5.0       & 48.8       & 0.175       & 0.122       & 2.1     \\ 
			\cmidrule(r){1-2} \cmidrule(r){3-6} \cmidrule(r){7-10}
			\multirow{ 3}{*}{(SCST, Co-att, $*$)}       & Ens$_{1}$(GAN$_{1}$)                       & 102.6     & 0.262     & \B{0.195}  & 9.9       & 44.8       & 0.172       & \B{0.129}   & \B{3.0} \\ 
			& Ens$_{2}$(GAN$_{2}$)                       & 115.1     & \B{0.277} & 0.194      & 7.0       & 48.3       & 0.176       & 0.127       & 2.7     \\ 
			& Ens$_{12}$(GAN$_{1}$,GAN$_{2}$)            & 113.2     & 0.274     & \B{0.195}  & 7.3       & 49.9       & 0.178       & \B{0.129}   & 2.6     \\ 
			\cmidrule(r){1-2} \cmidrule(r){3-6} \cmidrule(r){7-10}
			\multirow{ 3}{*}{(SCST, Joint-Emb, $*$)}    & Ens$_{3}$(GAN$_{3}$)                       & 109.8     & 0.270     & 0.193      & 8.5       & 48.5       & 0.175       & 0.127       & 2.8     \\ 
			& Ens$_{4}$(GAN$_{4}$)                       & 113.0     & 0.274     & 0.193      & 7.6       & 48.0       & 0.178       & 0.127       & 2.7     \\ 
			& Ens$_{34}$(GAN$_{3}$,GAN$_{4}$)            & 111.1     & 0.271     & 0.193      & 8.1       & \B{50.1}   & 0.177       & 0.127       & 2.8     \\ 
			\cmidrule(r){1-2} \cmidrule(r){3-6} \cmidrule(r){7-10}
			\multirow{ 3}{*}{(Gumbel $*$, Co-att, $*$)} & Ens$_{5}$(GAN$_{5}$)                       & 100.1     & 0.259     & 0.191      & \B{10.0}  & 43.1       & 0.170       & 0.127       & \B{3.0} \\ 
			& Ens$_{6}$(GAN$_{6}$)                       & 99.6      & 0.253     & 0.187      & 9.3       & 41.0       & 0.165       & 0.122       & 2.8     \\
			& Ens$_{7}$(GAN$_{7}$)                       & 100.2     & 0.254     & 0.180      & 7.8       & 38.9       & 0.164       & 0.113       & 2.3     \\ 
			& Ens$_{567}$(GAN$_{5}$,GAN$_{6}$,GAN$_{7}$) & 103.2     & 0.258     & 0.188      & 8.7       & 41.8       & 0.164       & 0.121       & 2.7     \\ 
			\cmidrule(r){1-2} \cmidrule(r){3-6} \cmidrule(r){7-10}
			(SCST+Gumbel Soft, Co-att, $*$)             & Ens$_{125}$(GAN$_{1}$,GAN$_{2}$,GAN$_{5}$) & 112.4     & 0.273     & 0.195  & 7.7       & 49.8       & \B{0.179}   & 0.129 & 2.7     \\ 
			\bottomrule
		\end{tabular}  
	} 
	\caption{Ensembling results for some GANs from Table~\ref{tab:GAN} for COCO and OOC sets. See \Tab{tab:ENSFULL} in \App{app:fulltables} for complete set of results including BLEU4 and ROUGEL.}
	\label{tab:ENS}
\end{table*}

\begin{figure*}[ht!]
	\centering
	\begin{subfigure}[t]{.3\linewidth}
		\centering
		\includegraphics[width=\linewidth]{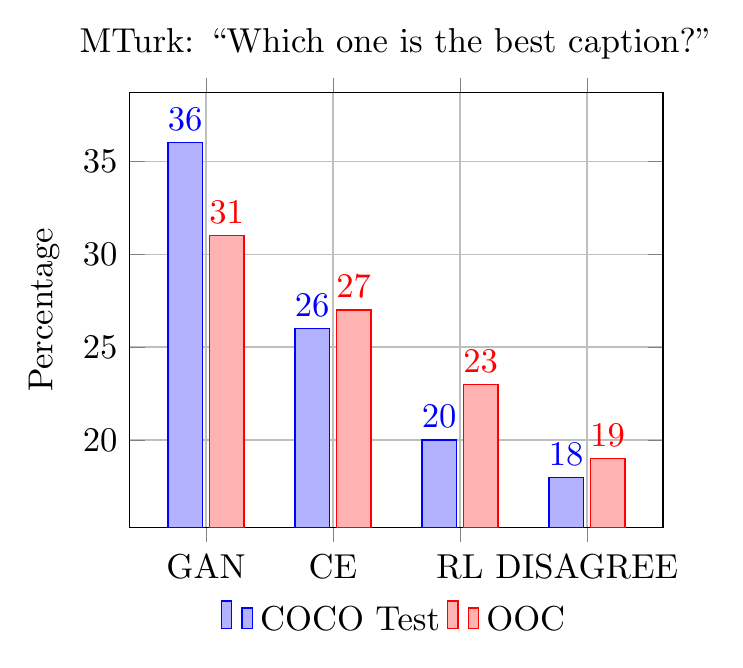}
		\caption{Best Caption}
	\end{subfigure}%
	\begin{subfigure}[t]{.33\linewidth}
		\centering
		\includegraphics[width=\linewidth]{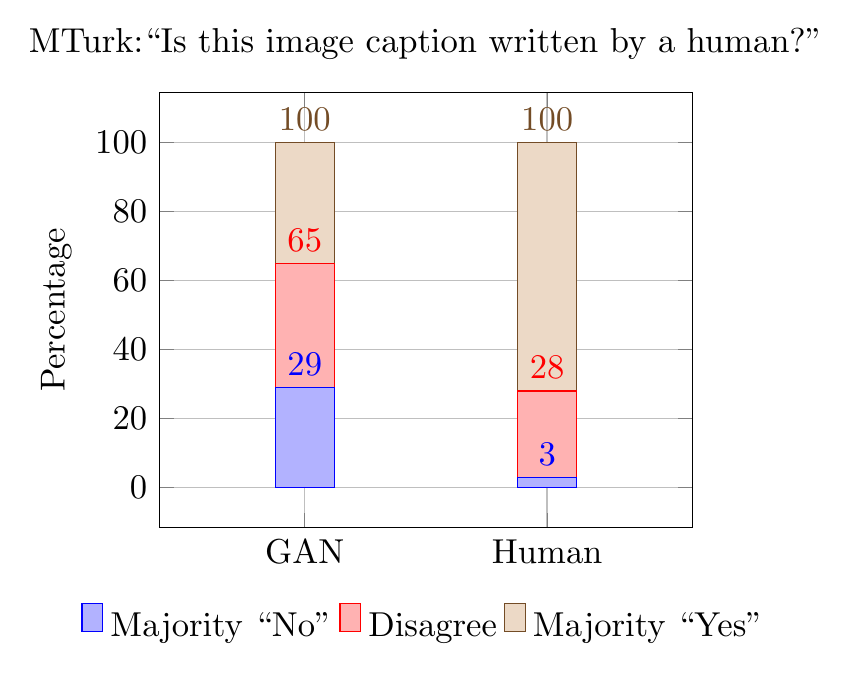}
		\caption{Turing Test}
	\end{subfigure}%
	\begin{subfigure}[t]{.3\linewidth}
		\centering
		\includegraphics[width=\linewidth]{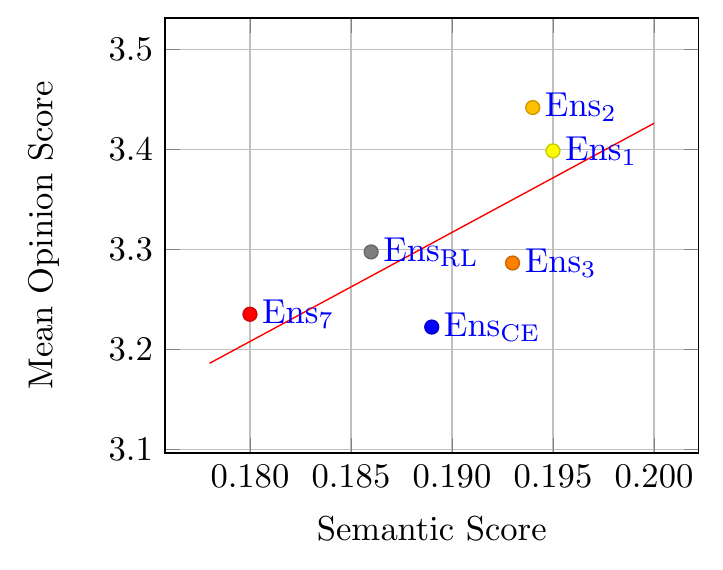}
		\caption{MOS COCO}
	\end{subfigure}
	\caption{Human evaluations of \Ens{\text{CE}}, \Ens{\text{RL}} and several GAN ensembles on COCO and OOC sets. (a) A distribution of preferences for the best caption determined by the majority of 5 human evaluators; here GAN label indicates the \Ens{2} model. (b) Turing test on detecting the human-written versus GAN-generated captions on COCO. We assign "yes/no' with at least 4 out of 5,  disagree otherwise. (c) Mean opinion score vs. Semantic score on COCO test images.}
	\label{fig:human}
\end{figure*}

\noindent \textbf{Experimental Results.} \Tab{tab:GAN} presents results for both COCO and OOC datasets for two discriminator architectures (ours Co-att, and baseline Joint-Emb) for all training algorithms (SCST, Gumbel ST, and Gumbel Soft). For reference, we also include results for non-GANs captioners: CE (trained only with cross entropy) and CIDEr-RL (pretrained with CE, followed by SCST to optimize CIDEr), as well as results from non-attentional models. As expected,  CIDEr-RL greatly improves the language metrics as compared to the CE model (from 101.6 to 116.1 CIDEr on COCO), but this also leads to a significant drop in  the vocabulary coverage (from 9.2\% to 5.1\% for COCO), indicating that the $n$-gram optimization can lead to vanilla sentences, discouraging style deviations from the ground truth captions.  In the table, \GAN{1} $,\dotsc,$ \GAN{4} denote the GAN-based models, where we use SCST training (with $\log(D)$ or $\log(D)+5\!\times\!\mathrm{CIDEr}$ rewards) with either Co-att or Joint-Emb discriminators; and \GAN{5} $,\dotsc,$ \GAN{7} are the models trained with the Gumbel relaxation. From our extensive experiments we observed that SCST provides significantly more stable training of the models and better results as compared to Gumbel approaches, which often become unstable beyond 15 epochs and underperform SCST GANs on many evaluation metrics (see also Section~\ref{app:expt} in Supplement for additional discussion on SCST vs. Gumbel).

It can noticed that SCST GAN models outperform CE and CIDEr-RL captioners on semantic score and vocabulary coverage for both COCO and OOC sets. The CIDEr regularization of SCST GAN additionally improves CIDEr and METEOR scores, and also results in the improvement of the semantic score (at the cost of some vocabulary coverage loss) as seen for \GAN{1} vs. \GAN{2} and \GAN{3} vs. \GAN{4}. We also see that SCST GANs using our Co-att discriminator outperform their Joint-Emb \cite{bodai} counterparts on \emph{every} metric except vocabulary coverage (for COCO).  We conclude that \GAN{2}, a CIDEr-regularized SCST with Co-att discriminator, is the model with the best overall performance on COCO and OOC sets.

For baselining, we also reproduced results from \cite{bodai} with non-attentional generators (same architecture as in \cite{bodai}).
Non-attentional models are behind in all metrics, except for vocabulary coverage on both datasets. Interestingly, Co-att discriminators still provide better semantic scores than Joint-Emb despite
non-attentional generators.


Figures~\ref{fig:sem_epoch_joint} and \ref{fig:sem_voc_joint} show the evolution of semantic scores and vocabulary coverage over the training epochs for \GAN{1}$,\dotsc$\GAN{4}, CE, CIDEr-RL and ground truth (GT) captions. Semantic scores increase steadily for all cost functions and discriminator architectures as the training sees more data.
In \Fig{fig:sem_epoch_joint}~(a), GAN models improve steadily over CE and RL, ultimately surpassing both of them mid-training.
Moreover, Co-att GANs achieve higher semantic scores across the epochs than Joint-Emb GANs.
For CIDEr-regularized SCST GANs, the same trend is observed but with a faster rate since the models start off worse than CE and RL.  For OOC in \Fig{fig:sem_epoch_joint}~(b), we see the same trend: Co-att GANs outperforming the other approaches. For COCO, GT semantic score is higher than the other models while the opposite is true for OOC.
This may be caused by the vocabulary mismatch between OOC and the combination of COCO and SBU. 
Figures~\ref{fig:sem_epoch_joint} and \ref{fig:sem_voc_joint} show that the semantic score improvement of GAN-trained models correlates well with the vocabulary coverage increase for both COCO and OOC.

\noindent \textbf{Ensemble Models.}
\Tab{tab:ENS} presents results for \emph{ensemble models}, where a caption is generated by first averaging the softmax scores from 4 different models before word selection. 
\Ens{\mathrm{CE}} and \Ens{\mathrm{RL}} ensemble CE and CIDEr-RL models. 
Similarly, \Ens{1}$,\dotsc$\Ens{7} ensemble models from \GAN{1}$,\dotsc$\GAN{7} respectively (\Ens{ijk} denotes an ensemble of \GAN{i}, \GAN{j}, and \GAN{k}).
As compared to individual models, the ensembles show improved results on \emph{all} metrics.

Ensembling SCST GANs provides the best results, reinforcing the conclusion that SCST is a superior method for a stable sequence GAN training.
For comparison, we also computed SPICE \cite{anderson2016spice} scores on COCO dataset: \Ens{\mathrm{CE}} 19.69, \Ens{12} 20.64 (GAN Co-attention) and \Ens{34} 20.46 (GAN Joint embedding from \cite{bodai}), showing that SCST GAN training additionally improves the SPICE metric.
Finally, we observe that underperformance of GANs over CIDEr-RL in terms of CIDEr is expected and explained by the fact that in GAN the objective is to make the sentences more descriptive and human-like, deviating from the vanilla ground truth captions, and this can potentially sacrifice the CIDEr performance.
The generated captions are evaluated using the proposed semantic score which showed a good correlation with human judgment; see \Fig{fig:human}~(c) for more details.

 \begin{figure*}[ht!]
  \centering
    \begin{subfigure}[t]{.4\textwidth}
      \centering
      \resizebox{0.95\linewidth}{!}{
        \begin{TAB}(c){|c|c|}{|c|c|c|c|}
          \includegraphics[width=4in, height=4in, keepaspectratio]{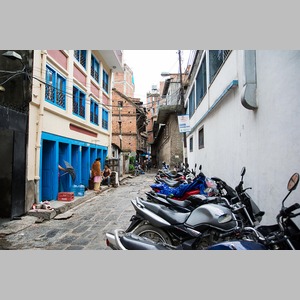} &
          \includegraphics[width=4in, height=4in, keepaspectratio]{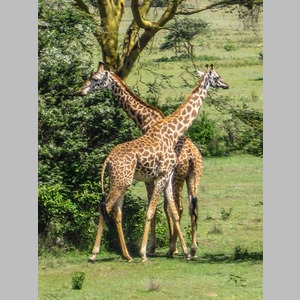} \\
          \parbox[][][c]{3.5in}{%
            GAN: a row of motorcycles parked on the side of a street \\
            CE: a row of motorcycles parked on a street \\
            RL: a group of motorcycles parked in front of a building \\
            GT: a bunch of motorcycles parked along the side of the street
          }
          &
          \parbox[][][c]{3.5in}{%
            GAN: two giraffes standing next to each other in a field \\
            CE: two giraffes standing in a field with trees in the background \\
            RL: a couple of giraffes standing next to each other \\
            GT: two giraffe standing next to each other on a grassy field
          }
          \\
          \includegraphics[width=4in, height=4in, keepaspectratio]{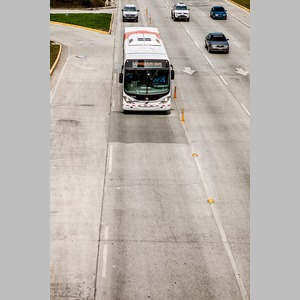} &
          \includegraphics[width=4in, height=4in, keepaspectratio]{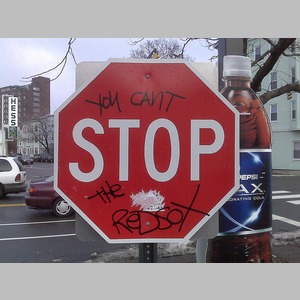} \\
          \parbox[][][c]{3.5in}{%
            GAN: a bus driving down a highway next to cars \\
            CE: a bus driving down a street next to a car \\
            RL: a bus driving down a highway with cars \\
            GT: a bus traveling on a freeway next to other traffic
          }
          &
          \parbox[][][c]{3.5in}{%
            GAN: a stop sign with graffiti on it next to a street \\
            CE: a stop sign with a sticker on it \\
            RL: a stop sign with a street on top of it \\
            GT: graffiti on a stop sign supporting the red sox
          }
        \end{TAB}%
      }
      \caption{COCO Test}
    \end{subfigure}%
    \begin{subfigure}[t]{.4\textwidth}
      \centering
      \resizebox{0.95\linewidth}{!}{          
        \begin{TAB}(c){|c|c|}{|c|c|c|c|}
          \includegraphics[width=4in, height=4in, keepaspectratio]{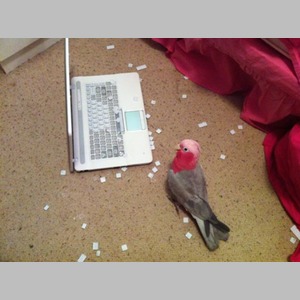} &
          \includegraphics[width=4in, height=4in, keepaspectratio]{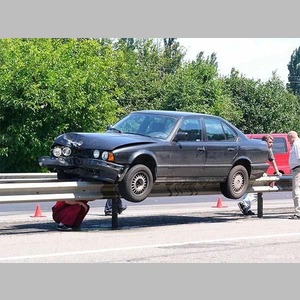} \\
          \parbox[][][c]{3.99in}{%
            GAN: a bird is sitting on the floor next to a laptop \\
            CE: a bird is sitting on the floor next to a laptop \\
            RL: a bird sitting on the floor next to a laptop \\
            GT: a bird on the ground with keys to the laptop are scattered on the floor
          }
          &
          \parbox[][][c]{3.99in}{%
            GAN: a black car parked on the side of the road \\
            CE: a black car parked on the side of the road \\
            RL: a black car parked on the side of a street \\
            GT: a black car has gotten lodged atop a silver metal barrier on a roadway
          }
          \\
          \includegraphics[width=4in, height=4in, keepaspectratio]{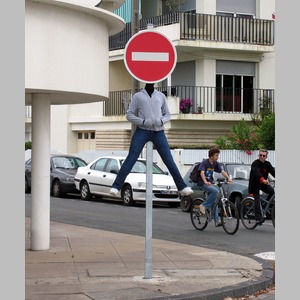} &
          \includegraphics[width=4in, height=4in, keepaspectratio]{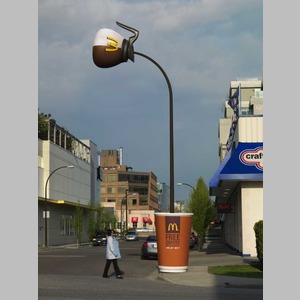} \\
          \parbox[][][c]{3.99in}{%
            GAN: a man riding a bike next to a stop sign \\
            CE: a man is walking down the street with a sign \\
            RL: a man riding a bike down a street with a stop sign \\
            GT: a person hangs from a street sign using the sign to cover their face
          }
          &
          \parbox[][][c]{3.99in}{%
            GAN: a person is standing in front of a coffee cup \\
            CE: a person is walking down a street with a cup \\
            RL: a cup of coffee sitting on top of a street \\
            GT: a person looks up at a street light that is designed like a pot of coffee
          }
        \end{TAB}%
      }
      \caption{OOC}
    \end{subfigure}%
  \caption{Examples of captions for our proposed model on COCO and OOC sets.}
  \label{fig:goodcap_joint}
\end{figure*}

\begin{figure}[!htb]
	\centering
	\includegraphics[width=\linewidth]{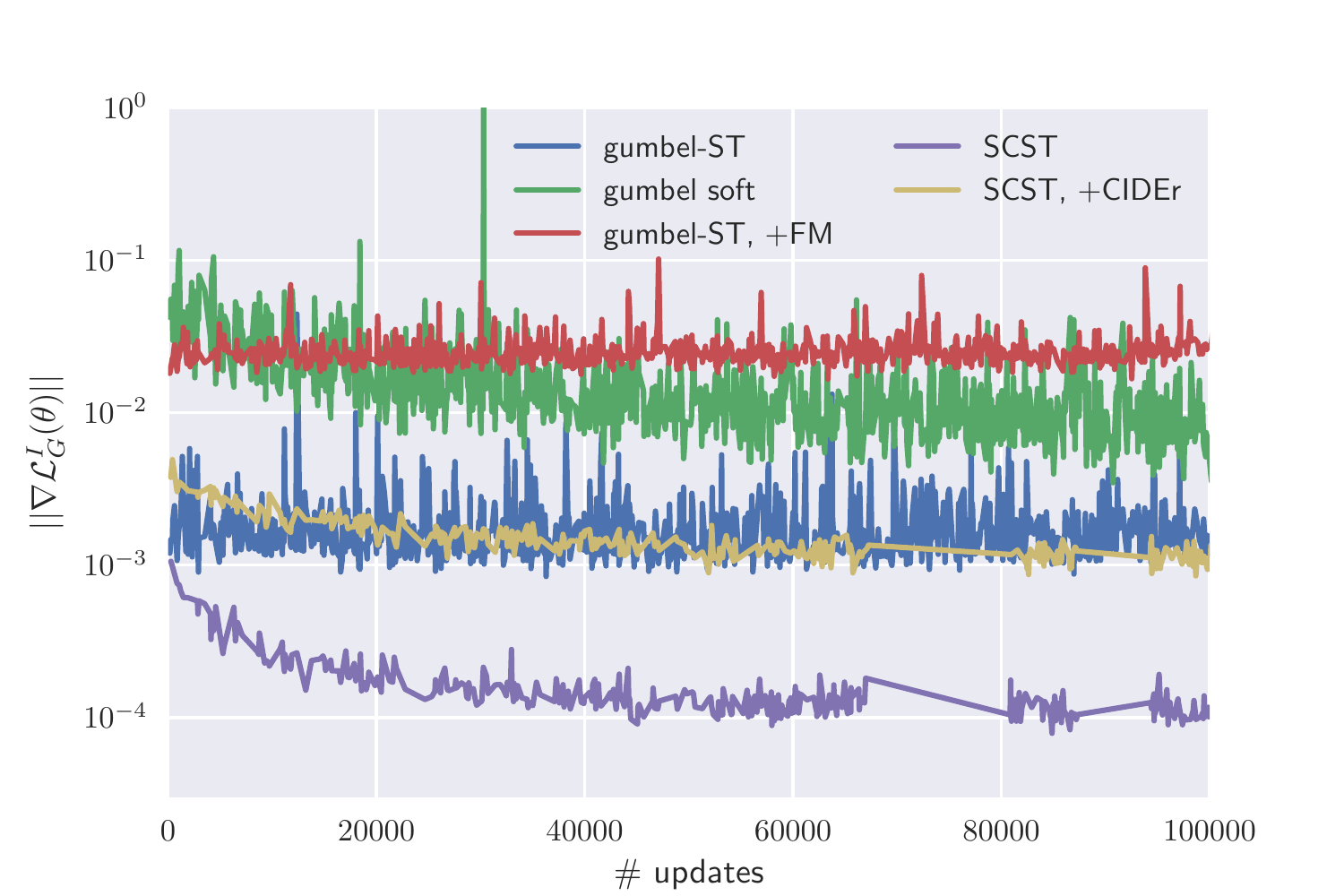}
	\caption{$L_2$ norm of the gradient with respect to the logits during training of $G_\theta$ with different training strategies.
	The plots show a minibatch-mean during the training; the variance of each curve gives a good idea of the gradient stability between minibatches.  We can see that SCST with pure discriminator reward has the lowest gradient norm.}
	\label{fig:gradient}
\end{figure}

\noindent \textbf{Gradient Analysis.} Throughout the extensive experiments, the SCST showed to be a more stable approach for training discrete GAN nodels, achieving better results compared to Gumbel relaxation approaches.
\Fig{fig:gradient} compares gradient behaviors during training for both techniques, showing that the SCST gradients have smaller average norm and variance across minibatches, confirming our conclusion.

\noindent \textbf{Human Evaluation.} To validate the benefits of the semantic score, we also evaluate the image/caption pairs generated by several GAN ensembles, \Ens{\text{CE}} and \Ens{\text{RL}} on Amazon MTurk. For a given image, 5 workers are asked to rate each caption on the scale 1 to 5 (from which we computed mean opinion score (MOS)) as well as to select the best caption overall (additional details are given in \App{app:human}).
\Fig{fig:human}~(a) shows that GAN ensemble \Ens{2} scored higher than CE and CIDEr-RL on a majority vote, confirming that GAN training significantly improves perceived quality of the captions as compared to a more vanilla CE or RL-based captions. \Fig{fig:human}~(b) gives Turing test results where the workers are asked if a given caption is human or machine-generated. Here, our GANs again performed well, demonstrating a good capacity at fooling humans. In \Fig{fig:human}~(c) we show that MOS of human evaluations correlates well with our semantic score (see \Tab{tab:MOS} in \App{app:human} for all scores). There is an overall trend (depicted with a red regression line for better visualization), where models that have higher semantic score are generally favored more by human evaluators. For example, Co-att SCST GANs \Ens{1} and \Ens{2} score the best semantic (resp. 0.195 and 0.194) and MOS scores (resp. 3.398 and 3.442) on COCO.
We can see that the semantic score is able to capture semantic alignments pertinent to humans, validating it as a viable alternative to automatic language metrics and a proxy to human evaluation.

Finally, in \Fig{fig:goodcap_joint} we present a few examples of the captions for COCO and OOC sets. As compared to the traditional COCO dataset, the OOC images are difficult and illustrate the challenge for the automatic captioning in such settings. The difficulty is not only to correctly recognize the objects in the scene but also to compose a proper description, which is challenging even for humans (see row denoted by GT), as it takes more words to describe such unusual images.

\section{Conclusion}

In conlusion, we summarize the main messages from our study:
\begin{enumerate*}[label={\arabic*)},font={\bfseries}]
\item SCST training for sequence GAN is a promissing new approach that outperforms the Gumbel relaxation in terms of stability of training and the overall performance.
\item The modeling part in the captioner is crucial for generalization to out-of-context: we demonstrate that the non-attention captioners and discriminators -- while still widely used -- fail at generalizing to out of context, hinting at a memorization of the training set. Attentive captioners and discriminators succeed at composing on unseen visual scenes, as  was demonstrated with our newly introduced OOC diagnostic set.
\item Human evaluation is still the \emph{gold standard} for assessing the quality of GAN captioning. We showed that the introduced semantic score correlates well with the human judgement and can be a valuable addition to the  existing evaluation toolbox for image captioning. 
\end{enumerate*}

\FloatBarrier

\bibliographystyle{ieee}
\bibliography{References}

\newpage
\appendix
{\onecolumn
	\centering
\section*{Adversarial Semantic Alignment for Improved Image Captions \\ (Supplementary Material)}
\vspace{1pt}
\subsection*{\normalfont
Pierre Dognin,  Igor Melnyk,  Youssef Mroueh,  Jerret Ross \& Tom Sercu \\
IBM Research, Yorktown Heights, NY}
\vspace{5pt}
}

\section{Semantic Score}\label{app:SemanticScore}

Semantic scores was first introduced int the context of image retrieval where it achieves state of the art performance \cite{assymetric}. Some examples of the properties of semantic scores are given in \Tab{tab:sem_sentence_table}.

\begin{table*}[!htb]
  \centering
  \resizebox{\textwidth}{!}{
    \begin{tabular}{ccrl}
      \toprule
      COCO validation image & Set & Semantic & Captions  \\
                            &        & Score    &          \\
      \midrule
      \multirow{25}{*}{\includegraphics[width=0.28\textwidth]{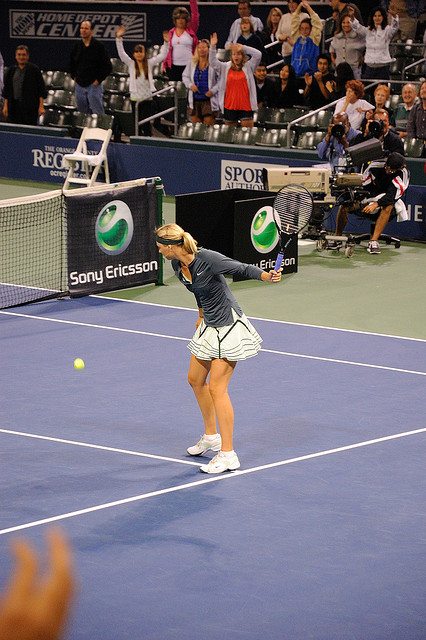}} &
       \multirow{4}{*}{Set A} & 0.181052  & female tennis player reaches back to swing at the ball \\
       & & 0.210224  & a woman on a court swinging a racket at a ball	\\
       & & 0.181592  & a woman in a gray top is playing tennis \\
       & & 0.251200  & the woman is playing tennis on the court \\	
       & & 0.145646  & a woman prepares to hit a tennis ball with a racket \\
       \cmidrule(r){2-4}
       & \multirow{5}{*}{Set B} & 0.008990  & a clear refrigerator is stocked up with food \\
       & & 0.005519  & a store freezer is shown with food inside \\	
       & & -0.014052 & a refrigerated display case is full of dairy groceries \\	
       & & 0.011076  & a close up of a commercial refrigerator with food \\
       & & -0.029001 & a large cooler with glass doors containing mostly dairy products \\
      \cmidrule(r){2-4}
       & \multirow{4}{*}{Set C} & 0.054441  & a {\color{red}giraffe} reaches back to swing at the ball \\
       & & 0.123822  & female tennis player reaches back to swing at the {\color{red}boat} \\
       & & 0.152860  & {\color{red}male} tennis player reaches back to swing at the ball \\
       & & 0.067289  & female {\color{red}football} player reaches back to swing at the ball \\
      \cmidrule(r){2-4}
      & \multirow{4}{*}{Set D} & 0.152860  & {\color{orange}male} tennis player reaches back to swing at the ball \\
      &  & 0.164755  & female tennis {\color{orange}fan} reaches back to swing at the ball \\	
      &  & 0.152524  & female tennis player {\color{orange}looks} back to swing at the ball \\
      &  & 0.100098  & female {\color{orange}flute} player reaches back to swing at the ball \\
       \cmidrule(r){2-4}
      & \multirow{5}{*}{Set E}  & 0.114010  & female tennis player swing ball \\
      &  &  0.031566  & female player swing ball \\
      &  &  0.084016  & tennis player swing ball \\
      &  &  0.115490  & tennis player ball \\	
      &  &  0.092226  & tennis player \\
      &  & -0.044019  & tennis \\	
      &  & -0.001948  & ball ball ball ball \\
     \bottomrule
    \end{tabular}
  }
  \caption{Semantic scores for various captions given an image from COCO validation set. Set A is composed of the  5 ground truth captions provided by COCO. Semantic scores are in between .14 and .25 for a possible range of [-1,1] being a cosine distance. Set B is made of captions from another randomly selected image in the validation set. The scores are clearly much worse (smaller) when captions do not match the image visual cues. Set C is a one-word modification set of the first caption in Set A. Semantic scores are all lower compared to the original caption. In Set C, we want to see if the metric is solely sensitive the main visual cues and if it can pick up subtle differences like gender.  Again, all the scores are still lower, even if closer to the original caption's score. In Set E, we are trying to break the metric by narrowing down to only factual words and objects. The combined knowledge of visual and text correlation penalize simplistic descriptive list of words. This does not imply that the metric cannot be fooled, but it seems resilient to obvious gaming like repeating words of some visual cues.}
  \label{tab:sem_sentence_table}
\end{table*}

\section{Experimental Results: Complete Tables}\label{app:fulltables}

We report here CIDEr, BLEU4, ROUGEL, METEOR, semantic scores, and vocabulary coverage for all models mentioned in this work, both COCO and OOC sets.
\Tab{tab:GANFULL} presents all GAN results as average ($\pm$~standard deviation) over 4 models with different random seeds.
\Tab{tab:ENSFULL} presents all our ensemble results. 

\begin{table*}[!htb]
  \caption{Collection of results for all models mentioned in this work. We provide commonly used CIDEr, BLEU4, ROUGEL, METEOR scores, as well as semantic scores, and percentage of vocabulary coverage for both COCO and OOC. Results are averaged from 4 models from independent trainings. We report mean and standard deviation for all metrics when available.}
  \label{tab:GANFULL}
  \centering
  \resizebox{\linewidth}{!}{
    \begin{tabular}{lrrrrrrrrrrrr}
      \toprule
      {} & \multicolumn{12}{l}{COCO Test Set} \\
      \cmidrule(r){2-13}
      {} & \multicolumn{2}{l}{CIDEr} & \multicolumn{2}{l}{BLEU4} & \multicolumn{2}{l}{ROUGEL} & \multicolumn{2}{l}{METEOR} & \multicolumn{2}{l}{Semantic} & \multicolumn{2}{l}{Vocabulary} \\
      {} & \multicolumn{2}{l}{}      & \multicolumn{2}{l}{}      & \multicolumn{2}{l}{}       & \multicolumn{2}{l}{}       & \multicolumn{2}{l}{Score}    & \multicolumn{2}{l}{Coverage}   \\
      \cmidrule(r){1-1} \cmidrule(r){2-13}
      CE                                                      & 101.6     & $\pm$0.4 & 0.312     & $\pm$.001 & 0.542     & $\pm$.001 & 0.260     & $\pm$.001 & 0.186     & $\pm$.001 & 9.2      & $\pm$0.1 \\ 
      CIDEr-RL                                                & \B{116.1} & $\pm$0.2 & \B{0.350} & $\pm$.003 & \B{0.562} & $\pm$.001 & 0.269     & $\pm$.000 & 0.184     & $\pm$.001 & 5.1      & $\pm$0.1 \\
      \cmidrule(r){1-1} \cmidrule(r){2-13}
      GAN$_{1}$(SCST, Co-att, $\log(D)$)                      & 97.5      & $\pm$0.8 & 0.294     & $\pm$.002 & 0.532     & $\pm$.001 & 0.256     & $\pm$.001 & 0.190     & $\pm$.000 & 11.0     & $\pm$0.1 \\
      GAN$_{2}$(SCST, Co-att, $\log(D)\!+\!5\times$CIDEr)     & 111.1     & $\pm$0.7 & 0.330     & $\pm$.004 & 0.555     & $\pm$.002 & \B{0.271} & $\pm$.002 & \B{0.192} & $\pm$.000 & 7.3      & $\pm$0.2 \\
      GAN$_{3}$(SCST, Joint-Emb, $\log(D)$)                   & 97.1      & $\pm$1.2 & 0.287     & $\pm$.005 & 0.530     & $\pm$.002 & 0.256     & $\pm$.002 & 0.188     & $\pm$.000 & \B{11.2} & $\pm$0.1 \\
      GAN$_{4}$(SCST, Joint-Emb, $\log(D)\!+\!5\times$CIDEr)  & 108.2     & $\pm$4.9 & 0.325     & $\pm$.017 & 0.551     & $\pm$.008 & 0.267     & $\pm$.004 & 0.190     & $\pm$.000 & 8.3      & $\pm$1.6 \\
      \cmidrule(r){1-1} \cmidrule(r){2-13}
      GAN$_{5}$(Gumbel Soft, Co-att, $\log(D)$)               & 93.6      & $\pm$3.3 & 0.282     & $\pm$.015 & 0.524     & $\pm$.007 & 0.253     & $\pm$.007 & 0.187     & $\pm$.002 & 11.1     & $\pm$1.2 \\ 
      GAN$_{6}$(Gumbel ST, Co-att, $\log(D)$)                 & 95.4      & $\pm$1.5 & 0.298     & $\pm$.009 & 0.531     & $\pm$.005 & 0.249     & $\pm$.004 & 0.184     & $\pm$.003 & 10.1     & $\pm$0.9 \\
      GAN$_{7}$(Gumbel ST, Co-att, $\log(D)+$FM)              & 92.1      & $\pm$5.4 & 0.289     & $\pm$.020 & 0.523     & $\pm$.015 & 0.243     & $\pm$.011 & 0.175     & $\pm$.006 & 8.6      & $\pm$0.8 \\
      \cmidrule(r){1-1} \cmidrule(r){2-13}
      G-GAN \cite{bodai} from Table 1                         & 79.5      &          & 0.207     &           & 0.475     &           & 0.224     &           & --        &           & --       &          \\
      \cmidrule(r){1-1} \cmidrule(r){2-13}
      \FC{CE}   -- \FC{} for non-attentional models           & 87.6      & $\pm$1.2 & 0.275     & $\pm$.003 & 0.516     & $\pm$.003 & 0.242     & $\pm$.001 & 0.175     & $\pm$.002 & 9.9      & $\pm$0.8 \\
      \FC{CIDEr-RL}                                           & 100.4     & $\pm$7.9 & 0.305     & $\pm$.018 & 0.536     & $\pm$.010 & 0.253     & $\pm$.006 & 0.173     & $\pm$.002 & 6.8      & $\pm$1.4 \\
      \cmidrule(r){1-1} \cmidrule(r){2-13}
      \FC{GAN$_{1}$}(SCST, Co-att, $\log(D)$)                 & 89.7      & $\pm$0.9 & 0.276     & $\pm$.000 & 0.518     & $\pm$.001 & 0.246     & $\pm$.001 & 0.184     & $\pm$.001 & 13.2     & $\pm$0.2 \\
      \FC{GAN$_{2}$}(SCST, Co-att, $\log(D)+5\times$CIDEr)    & 103.1     & $\pm$0.5 & 0.311     & $\pm$.003 & 0.542     & $\pm$.001 & 0.261     & $\pm$.001 & 0.183     & $\pm$.001 & 7.1      & $\pm$0.2 \\
      \FC{GAN$_{3}$}(SCST, Joint-Emb, $\log(D)$)              & 90.7      & $\pm$0.1 & 0.277     & $\pm$.002 & 0.520     & $\pm$.000 & 0.248     & $\pm$.001 & 0.181     & $\pm$.001 & 12.9     & $\pm$0.1 \\
      \FC{GAN$_{4}$}(SCST, Joint-Emb, $\log(D)+5\times$CIDEr) & 102.7     & $\pm$0.4 & 0.315     & $\pm$.000 & 0.542     & $\pm$.000 & 0.260     & $\pm$.001 & 0.182     & $\pm$.001 & 7.7      & $\pm$0.1 \\
      \midrule
      {} & \multicolumn{12}{l}{OOC (Out of Context)}  \\
      \cmidrule(r){2-13}
      {} & \multicolumn{2}{l}{CIDEr} & \multicolumn{2}{l}{BLEU4} & \multicolumn{2}{l}{ROUGEL} & \multicolumn{2}{l}{METEOR} & \multicolumn{2}{l}{Semantic} & \multicolumn{2}{l}{Vocabulary} \\
      {} & \multicolumn{2}{l}{}      & \multicolumn{2}{l}{}      & \multicolumn{2}{l}{}       & \multicolumn{2}{l}{}       & \multicolumn{2}{l}{Score}    & \multicolumn{2}{l}{Coverage}   \\
      \cmidrule(r){1-1} \cmidrule(r){2-13}
      CE                                                      & 42.2     & $\pm$0.6 & 0.168     & $\pm$.005 & 0.413     & $\pm$.003 & 0.169     & $\pm$.001 & 0.118     & $\pm$.001 & 2.8     & $\pm$0.1 \\
      CIDEr-RL                                                & 45.0     & $\pm$0.6 & 0.177     & $\pm$.002 & 0.417     & $\pm$.004 & 0.170     & $\pm$.003 & 0.117     & $\pm$.002 & 2.1     & $\pm$0.0 \\
      \cmidrule(r){1-1} \cmidrule(r){2-13}
      GAN$_{1}$(SCST, Co-att, $\log(D)$)                      & 41.0     & $\pm$1.6 & 0.161     & $\pm$.013 & 0.406     & $\pm$.006 & 0.168     & $\pm$.003 & \B{0.124} & $\pm$.000 & 3.2     & $\pm$0.1 \\
      GAN$_{2}$(SCST, Co-att, $\log(D)+5\times$CIDEr)         & \B{45.8} & $\pm$0.9 & 0.179     & $\pm$.014 & 0.417     & $\pm$.005 & \B{0.173} & $\pm$.001 & 0.122     & $\pm$.002 & 2.8     & $\pm$0.1 \\
      GAN$_{3}$(SCST, Joint-Emb, $\log(D)$)                   & 41.8     & $\pm$1.6 & 0.162     & $\pm$.006 & 0.404     & $\pm$.006 & 0.167     & $\pm$.002 & 0.122     & $\pm$.001 & \B{3.3} & $\pm$0.0 \\
      GAN$_{4}$(SCST, Joint-Emb, $\log(D)+5\times$CIDEr)      & 45.4     & $\pm$1.4 & \B{0.180} & $\pm$.011 & \B{0.418} & $\pm$.005 & \B{0.173} & $\pm$.002 & 0.122     & $\pm$.003 & 2.8     & $\pm$0.2 \\
      \cmidrule(r){1-1} \cmidrule(r){2-13}
      GAN$_{5}$(gumbel soft, Co-att, $\log(D)$)               & 38.3     & $\pm$3.7 & 0.154     & $\pm$.020 & 0.406     & $\pm$.006 & 0.164     & $\pm$.006 & 0.121     & $\pm$.004 & \B{3.3} & $\pm$0.3 \\
      GAN$_{6}$(gumbel-ST, Co-att, $\log(D)$)                 & 38.5     & $\pm$1.9 & 0.148     & $\pm$.005 & 0.407     & $\pm$.004 & 0.161     & $\pm$.005 & 0.116     & $\pm$.004 & 3.0     & $\pm$0.2 \\
      GAN$_{7}$(gumbel-ST, Co-att, $\log(D)$+FM)              & 36.8     & $\pm$2.3 & 0.154     & $\pm$.012 & 0.396     & $\pm$.009 & 0.157     & $\pm$.006 & 0.110     & $\pm$.005 & 2.5     & $\pm$0.2 \\
      \cmidrule(r){1-1} \cmidrule(r){2-13}
      \FC{CE}                                                 & 32.0     & $\pm$0.4 & 0.132     & $\pm$.007 & 0.392     & $\pm$.002 & 0.152     & $\pm$.002 & 0.103     & $\pm$.002 & 2.6     & $\pm$.1  \\
      \FC{CIDEr-RL}                                           & 33.4     & $\pm$1.4 & 0.145     & $\pm$.009 & 0.394     & $\pm$.006 & 0.154     & $\pm$.003 & 0.101     & $\pm$.003 & 2.1     & $\pm$.2  \\
      \cmidrule(r){1-1} \cmidrule(r){2-13}
      \FC{GAN$_{1}$}(SCST, Co-att, $\log(D)$)                 & 30.8     & $\pm$1.0 & 0.127     & $\pm$.001 & 0.383     & $\pm$.006 & 0.155     & $\pm$.003 & 0.111     & $\pm$.001 & 3.4     & $\pm$0.1 \\
      \FC{GAN$_{2}$}(SCST, Co-att, $\log(D)+5\times$CIDEr)    & 33.7     & $\pm$1.9 & 0.145     & $\pm$.011 & 0.391     & $\pm$.004 & 0.157     & $\pm$.001 & 0.108     & $\pm$.001 & 2.7     & $\pm$0.1 \\
      \FC{GAN$_{3}$}(SCST, Joint-Emb, $\log(D)$)              & 30.8     & $\pm$2.1 & 0.126     & $\pm$.009 & 0.380     & $\pm$.004 & 0.153     & $\pm$.002 & 0.108     & $\pm$.001 & 3.5     & $\pm$0.1 \\
      \FC{GAN$_{4}$}(SCST, Joint-Emb, $\log(D)+5\times$CIDEr) & 33.3     & $\pm$2.4 & 0.144     & $\pm$.016 & 0.391     & $\pm$.006 & 0.157     & $\pm$.004 & 0.106     & $\pm$.000 & 2.7     & $\pm$0.1 \\      
      \bottomrule
    \end{tabular}    
  } 
\end{table*}

\begin{table*}[!htb]
  \caption{Collection of ensembling results for GAN models from Table~\ref{tab:GAN}. We provide commonly used CIDEr, BLEU4, ROUGEL, METEOR scores, as well as semantic scores, and percentage of vocabulary coverage for both COCO and OOC.}
  \label{tab:ENSFULL}
  \centering
  \resizebox{\textwidth}{!}{
    \begin{tabular}{llrrrrrrrrrrrrrrrrrr}
      \toprule
      {} & {} & \multicolumn{12}{l}{COCO Test Set} \\ 
      \cmidrule(r){3-14}
      {} & {} & \multicolumn{2}{l}{CIDEr} & \multicolumn{2}{l}{BLEU4} & \multicolumn{2}{l}{ROUGEL} & \multicolumn{2}{l}{METEOR} & \multicolumn{2}{l}{Semantic} & \multicolumn{2}{l}{Vocabulary} \\
      {} & {} & \multicolumn{2}{l}{}      & \multicolumn{2}{l}{}      & \multicolumn{2}{l}{}       & \multicolumn{2}{l}{}       & \multicolumn{2}{l}{Score}    & \multicolumn{2}{l}{Coverage}   \\ 
      \cmidrule(r){1-2} \cmidrule(r){3-14}
      \multirow{ 2}{*}{(CE and RL Baselines)}     & Ens$_{\text{CE}}$(CE)                      & 105.8     &  & 0.327     &  & 0.553     &  & 0.266     &  & 0.189     &  & 8.4      & \\ 
                                                  & Ens$_{\text{RL}}$(CIDEr-RL)                & \B{118.9} &  & \B{0.359} &  & \B{0.568} &  & 0.273     &  & 0.186     &  & 5.0      & \\ 
      \cmidrule(r){1-2} \cmidrule(r){3-14}
      \multirow{ 3}{*}{(SCST, Co-att, $*$)}       & Ens$_{1}$(GAN$_{1}$)                       & 102.6     &  & 0.314     &  & 0.543     &  & 0.262     &  & \B{0.195} &  & 9.9      & \\ 
                                                  & Ens$_{2}$(GAN$_{2}$)                       & 115.1     &  & 0.347     &  & 0.566     &  & \B{0.277} &  & 0.194     &  & 7.0      & \\ 
                                                  & Ens$_{12}$(GAN$_{1}$,GAN$_{2}$)            & 113.2     &  & 0.344     &  & 0.564     &  & 0.274     &  & \B{0.195} &  & 7.3      & \\ 
      \cmidrule(r){1-2} \cmidrule(r){3-14}
      \multirow{ 3}{*}{(SCST, Joint-Emb, $*$)}    & Ens$_{3}$(GAN$_{3}$)                       & 109.8     &  & 0.331     &  & 0.556     &  & 0.270     &  & 0.193     &  & 8.5      & \\ 
                                                  & Ens$_{4}$(GAN$_{4}$)                       & 113.0     &  & 0.343     &  & 0.562     &  & 0.274     &  & 0.193     &  & 7.6      & \\ 
                                                  & Ens$_{34}$(GAN$_{3}$,GAN$_{4}$)            & 111.1     &  & 0.335     &  & 0.558     &  & 0.271     &  & 0.193     &  & 8.1      & \\ 
      \cmidrule(r){1-2} \cmidrule(r){3-14}
      \multirow{ 3}{*}{(Gumbel $*$, Co-att, $*$)} & Ens$_{5}$(GAN$_{5}$)                       & 100.1     &  & 0.307     &  & 0.538     &  & 0.259     &  & 0.191     &  & \B{10.0} & \\ 
                                                  & Ens$_{6}$(GAN$_{6}$)                       & 99.6      &  & 0.313     &  & 0.541     &  & 0.253     &  & 0.187     &  & 9.3      & \\ 
                                                  & Ens$_{7}$(GAN$_{7}$)                       & 100.2     &  & 0.321     &  & 0.543     &  & 0.254     &  & 0.180     &  & 7.8      & \\ 
                                                  & Ens$_{567}$(GAN$_{5}$,GAN$_{6}$,GAN$_{7}$) & 103.2     &  & 0.327     &  & 0.550     &  & 0.258     &  & 0.188     &  & 8.7      & \\ 
      \cmidrule(r){1-2} \cmidrule(r){3-14}
      (SCST+Gumbel Soft, Co-att, $*$)             & Ens$_{125}$(GAN$_{1}$,GAN$_{2}$,GAN$_{5}$) & 112.4     &  & 0.343     &  & 0.562     &  & 0.273     &  & \B{0.195} &  & 7.7      & \\ 
      \midrule
      {} & {} & \multicolumn{12}{l}{OOC (Out of Context} \\ 
      \cmidrule(r){3-14}
      {} & {} & \multicolumn{2}{l}{CIDEr} & \multicolumn{2}{l}{BLEU4} & \multicolumn{2}{l}{ROUGEL} & \multicolumn{2}{l}{METEOR} & \multicolumn{2}{l}{Semantic} & \multicolumn{2}{l}{Vocabulary} \\
      {} & {} & \multicolumn{2}{l}{}      & \multicolumn{2}{l}{}      & \multicolumn{2}{l}{}       & \multicolumn{2}{l}{}       & \multicolumn{2}{l}{Score}    & \multicolumn{2}{l}{Coverage}   \\ 
      \cmidrule(r){1-2} \cmidrule(r){3-14}
      \multirow{ 2}{*}{(CE and RL Baselines)}     & Ens$_{\text{CE}}$(CE)                        & 44.8     &  & 0.177     &  & 0.423     &  & 0.172     &  & 0.122     &  & 2.6     & \\
                                                  & Ens$_{\text{RL}}$(RL)                        & 48.8     &  & \B{0.198} &  & 0.427     &  & 0.175     &  & 0.122     &  & 2.1     & \\
      \cmidrule(r){1-2} \cmidrule(r){3-14}
      \multirow{ 3}{*}{(SCST, Co-att, $*$)}       & Ens$_{1}$(GAN$_{1}$)                         & 44.8     &  & 0.175     &  & 0.422     &  & 0.172     &  & \B{0.129} &  & \B{3.0} & \\
                                                  & Ens$_{2}$(GAN$_{2}$)                         & 48.3     &  & 0.189     &  & 0.429     &  & 0.176     &  & 0.127     &  & 2.7     & \\
                                                  & Ens$_{12}$(GAN$_{1}$$ + $$4\times$GAN$_{2}$) & 49.9     &  & 0.197     &  & \B{0.437} &  & 0.178     &  & \B{0.129} &  & 2.6     & \\
      \cmidrule(r){1-2} \cmidrule(r){3-14}
      \multirow{ 3}{*}{(SCST, Joint-Emb, $*$)}    & Ens$_{3}$(GAN$_{3}$)                         & 48.5     &  & \B{0.198} &  & 0.429     &  & 0.175     &  & 0.127     &  & 2.8     & \\
                                                  & Ens$_{4}$(GAN$_{4}$)                         & 48.0     &  & 0.185     &  & 0.432     &  & 0.178     &  & 0.127     &  & 2.7     & \\
                                                  & Ens$_{34}$(GAN$_{3}$$ + $$4\times$GAN$_{4}$) & \B{50.1} &  & 0.195     &  & 0.435     &  & 0.177     &  & 0.127     &  & 2.8     & \\
      \cmidrule(r){1-2} \cmidrule(r){3-14}
      \multirow{ 3}{*}{(Gumbel $*$, Co-att, $*$)} & Ens$_{5}$(GAN$_{5}$)                         & 43.1     &  & 0.169     &  & 0.420     &  & 0.170     &  & 0.127     &  & \B{3.0} & \\
                                                  & Ens$_{6}$(GAN$_{6}$)                         & 41.0     &  & 0.155     &  & 0.420     &  & 0.165     &  & 0.122     &  & 2.8     & \\
                                                  & Ens$_{7}$(GAN$_{7}$)                         & 38.9     &  & 0.166     &  & 0.413     &  & 0.164     &  & 0.113     &  & 2.3     & \\
                                                  & Ens$_{567}$(GAN$_{5}$,GAN$_{6}$,GAN$_{7}$)   & 41.8     &  & 0.167     &  & 0.418     &  & 0.164     &  & 0.121     &  & 2.7     & \\
      \cmidrule(r){1-2} \cmidrule(r){3-14}
      (SCST+Gumbel Soft, Co-att, $*$)             & Ens$_{125}$(GAN$_{1}$,GAN$_{2}$,GAN$_{5}$)   & 49.8     &  & \B{0.198} &  & 0.436     &  & \B{0.179} &  & \B{0.129} &  & 2.7     & \\
      \bottomrule
    \end{tabular}  
  } 
\end{table*}

\section{Semantic and Discriminator Scores Correlation over Training Epochs}\label{app:SemvsDisc}

\begin{figure*}[!htb]
  \centering
  \resizebox{0.8\linewidth}{!}{
    \includegraphics[width=0.9\textwidth]{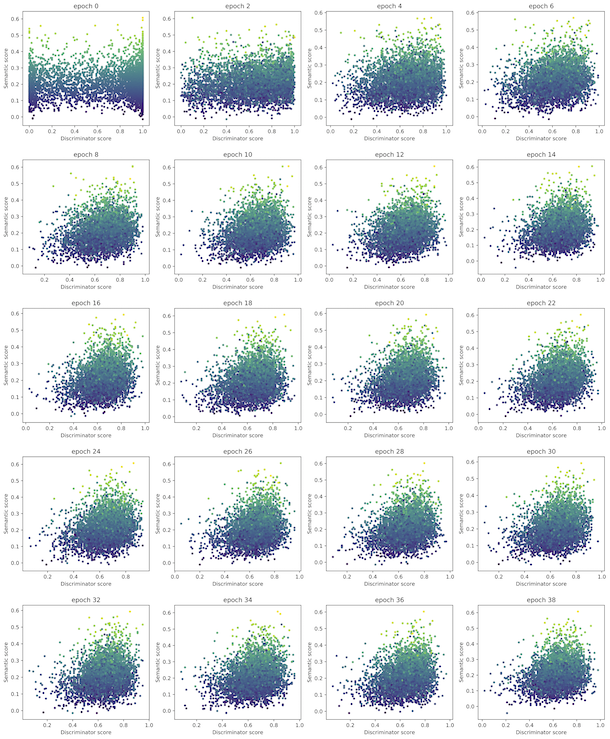}
  }
  \caption{Semantic vs. Discriminator scores across 40 training epochs for ground truth captions using the joint embedding discriminator \cite{bodai}.}
  \label{fig:scatterBoDaiGT}  
\end{figure*}

We are interested in the correlation between the semantic scores and discriminator scores of image captions as well as its evolution along the process of SCST GAN training.
We provide scatter plots for the Joint-Embedding discriminator \cite{bodai} across training in \Fig{fig:scatterBoDaiGT}. This GAN model was trained over 40 epochs with a discriminator pretrained on 15 epochs of data.

We compare semantic scores and discriminator scores over training epochs given the ground truth (GT) caption for each image in the COCO Test set (5K images). Each GT caption being fixed, we can observe the evolution of the semantic and discriminator score without any other effects. \Fig{fig:scatterBoDaiGT} show the semantic score, discriminator score pairs for each image (one point per image) for the joint embedding discriminator.
Since the GT captions are fixed, the semantic scores will be identical across epochs. From the first epoch, the joint embedding discriminator provides a wide range of scores with most scores close to the 0.0 and 1.0 min/max values.
Quickly the points cluster into a 'sail' like shape in the lower right corner, away from the min/max edges.
The color assigned to each point is directly linked to the semantic scores assigned at the first epoch of training. You can therefore have a small visual cue of the movement of these points from epoch to epoch and witness the discriminator learning how to distinguish real and fake captions.

\section{Human Evaluation}\label{app:human}
In this section we present the details of our evaluation protocol for our captioning models on Amazon MTurk.
All images are presented to 5 workers and aggregated in mean opinion score (MOS) or majority vote.

\paragraph{Turing Test.} In this setting we give human evaluators an image with a sentence either generated from our GAN captioning models or the ground truth. We ask them whether the sentence is human generated or machine generated. Exact instructions are: "Is this image caption written by a human? Yes/No. The caption could be written by a human or by a computer, more or less 50-50 chance."

\begin{figure*}[htb]
\centering
\includegraphics[width=0.8\textwidth]{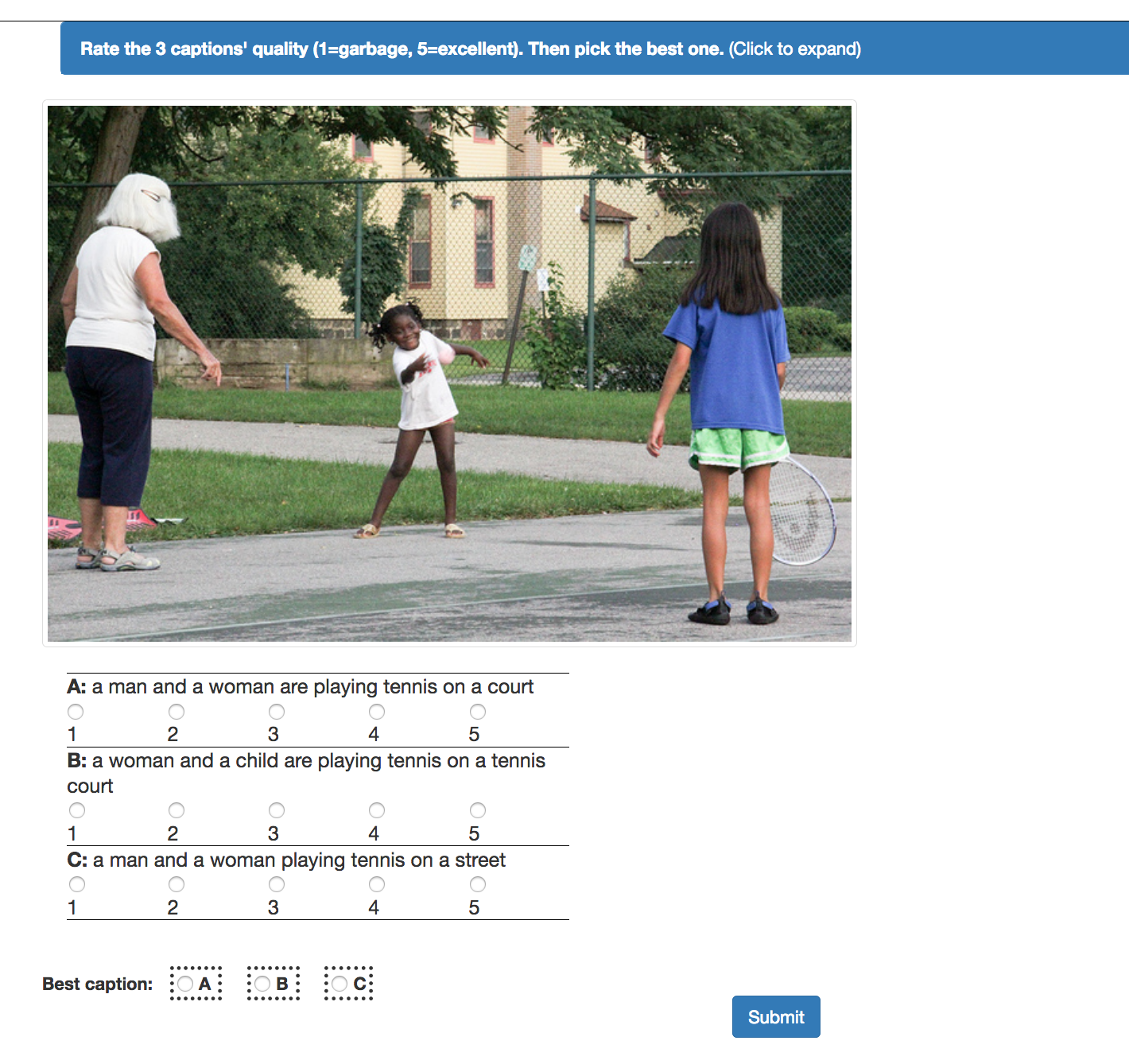}
\caption{The interface of "Fine Grained Evaluation".
}
\label{fig:turk}
\end{figure*}

\paragraph{Fine Grained Evaluation and Model Comparison.} In this experiment we give human evaluators an image and a set of 3 captions: Generated by CE trained model, SCST CIDEr trained model, and a GAN model. We ask them to rate each sentence on a scale of one to five.
After rating, the worker chooses the caption he/she thinks is best at describing the image.
In Section~\ref{sec:expres}, we provide results for Mean Opinion Score and Majority vote based of this interface (see \Fig{fig:turk}) and \Tab{tab:MOS}.

\begin{table*}[!htb]
  \caption{MOS and semantic scores collected from Amazon MTurk.}
  \label{tab:MOS}
  \centering
  \resizebox{0.8\linewidth}{!}{    
    \begin{tabular}{lcccc}
      \toprule
     
      {}                                                     & \multicolumn{2}{l}{COCO Test} & \multicolumn{2}{l}{OOC}            \\
      \cmidrule(r){2-3} \cmidrule(r){4-5}
      {}                                                     & Semantic Score                & MOS       & Semantic Score & MOS   \\
      \cmidrule(r){1-1}  \cmidrule(r){2-2} \cmidrule(r){3-3} \cmidrule(r){4-4} \cmidrule(r){5-5} 
      \Ens{\mathrm{CE}}(CE)                                  & 0.189                         & 3.222     & 0.122          & 3.065 \\
      \Ens{\mathrm{RL}}(CIDEr-RL)                            & 0.186                         & 3.297     & 0.122          & 3.097 \\
      \cmidrule(r){1-1}  \cmidrule(r){2-2} \cmidrule(r){3-3} \cmidrule(r){4-4} \cmidrule(r){5-5} 
      \Ens{1}(SCST, Co-att, $\log(D)$)                       & \B{0.195}                     & 3.398     &    --          &  --   \\
      \Ens{2}(SCST, Co-att, $\log(D)+5\times\mathrm{CIDEr}$) & 0.194                         & \B{3.442} & 0.127          & 3.107 \\
      \Ens{3}(SCST, Joint-Emb, $\log(D)$)                    & 0.193                         & 3.286     &    --          &  --   \\
      \cmidrule(r){1-1}  \cmidrule(r){2-2} \cmidrule(r){3-3} \cmidrule(r){4-4} \cmidrule(r){5-5} 
      \Ens{5}(Gumbel Soft, Co-Att,$\log(D)$)                 & 0.191                         & 3.138     &    --          &  --   \\
      \Ens{7}(Gumbel ST, Co-Att, $\log(D)+\mathrm{FM}$)      & 0.180                         & 3.235     &    --          &  --   \\
      \bottomrule
    \end{tabular}
  }
\end{table*}

\section{Experimental Protocol SCST vs. Gumbel}\label{app:expt}

\begin{figure*}[ht!]
  \centering
  \resizebox{1.0\linewidth}{!}{
    \includegraphics[width=1.0\textwidth]{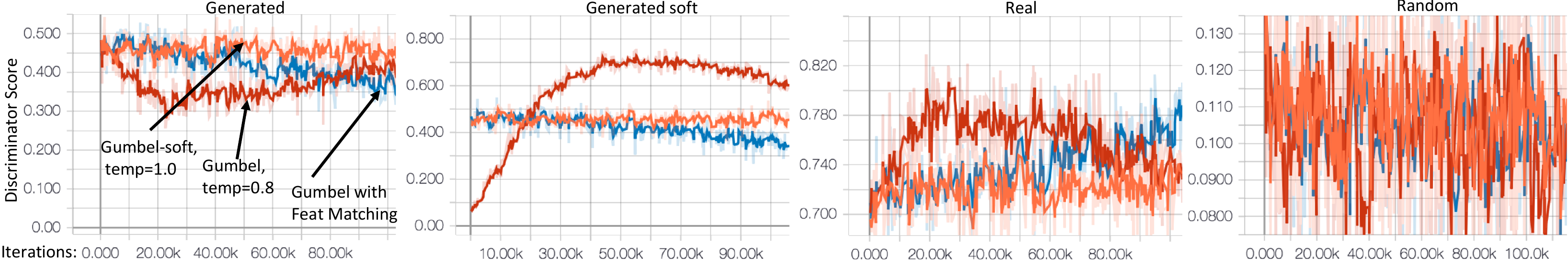}
  }
  \caption{Discrimator scores across different training Gumbel Methods.}
  \label{fig:SemVsDisc}  
\end{figure*}

In \Fig{fig:SemVsDisc}, we show that all our Gumbel Methods trained effectively. We plot the Discriminator scores (averaged over minibatch) during training with the 3 reported Gumbel models. Generated sentences get roughly 0.5, random sentences around 0.1, real sentences around 0.75. Hence, the Discriminator can correctly distinguish real from random, and generated sentences. This indicates a healthy training of all Gumbel Methods.

\section{Examples of Generated Captions}\label{app:examples}
\begin{figure*}[htb]
  \centering
  \begin{subfigure}[t]{.9\linewidth}
    \centering
    \resizebox{0.9\linewidth}{!}{        
      \begin{TAB}(c){|c|c|}{|c|c|c|c|}
        \includegraphics[width=4in, height=4in, keepaspectratio]{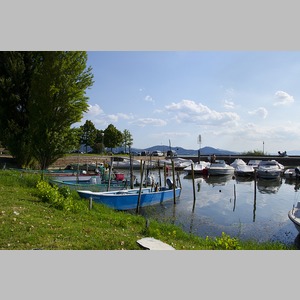} &
        \includegraphics[width=4in, height=4in, keepaspectratio]{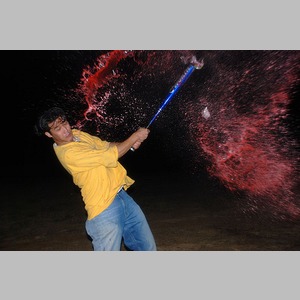} \\
        \parbox[][][c]{3.5in}{%
          GAN: a group of boats are docked in a harbor \\
          CE: a group of boats sitting in the water \\
          RL: a group of boats sitting in the water \\
          GT: some boats parked in the water at a dock
        }
        &
        \parbox[][][c]{3.5in}{%
          GAN: a man holding a baseball bat at night \\
          CE: a man is holding a bat in a dark \\
          RL: a man holding a baseball bat at a ball \\
          GT: a boy in yellow shirt swinging a baseball bat
        }
        \\
        \includegraphics[width=4in, height=4in, keepaspectratio]{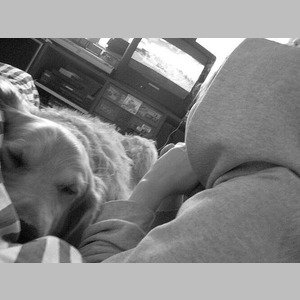} &
        \includegraphics[width=4in, height=4in, keepaspectratio]{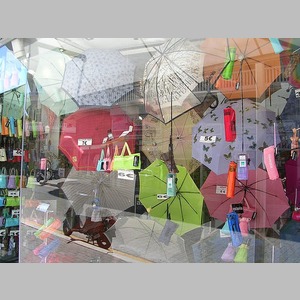} \\
        \parbox[][][c]{3.5in}{%
          GAN: a dog laying down on a person 's lap \\
          CE: a dog is sleeping on a couch with a person \\
          RL: a dog laying on a couch with a person \\
          GT: a dog that is laying next to another person
        }
        &
        \parbox[][][c]{3.5in}{%
          GAN: a bunch of umbrellas hanging from a wall in a store \\
          CE: a bunch of umbrellas that are hanging from a wall \\
          RL: a group of umbrellas hanging from a store \\
          GT: a bunch of umbrellas that are behind a glass
        }
      \end{TAB}
    }
  \end{subfigure}
  \caption{Cherry-picked examples on the COCO validation set.}
  \label{fig:goodcoco}
\end{figure*}

\begin{figure*}[htb]
  \centering
  \begin{subfigure}[t]{.9\linewidth}
    \centering
    \resizebox{0.9\linewidth}{!}{         
      \begin{TAB}(c){|c|c|}{|c|c|c|c|}
        \includegraphics[width=4in, height=4in, keepaspectratio]{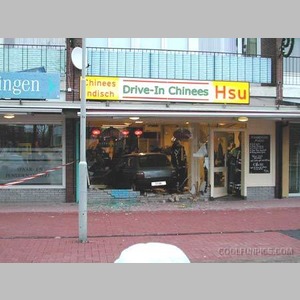} &
        \includegraphics[width=4in, height=4in, keepaspectratio]{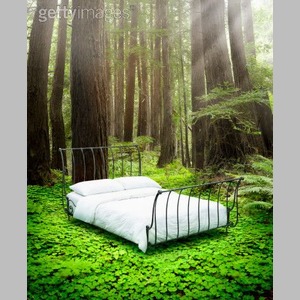} \\
        \parbox[][][c]{3.5in}{%
          GAN: a store front with a car parked in front of it \\
          CE: a store with a sign on the side of it \\
          RL: a building with a sign on the side of it \\
          GT: a car has crashed into the store front of a chinese restaurant
        }
        &
        \parbox[][][c]{3.5in}{%
          GAN: a bed sitting in the middle of a forest \\
          CE: a bed with a green blanket on it \\
          RL: a bed in a forest with a table \\
          GT: a bed lies on top of a clover field in a forest
        }
        \\
        \includegraphics[width=4in, height=4in, keepaspectratio]{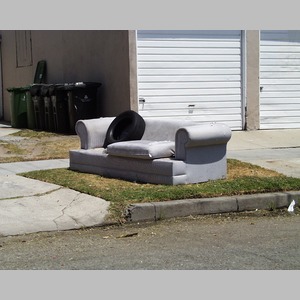} &
        \includegraphics[width=4in, height=4in, keepaspectratio]{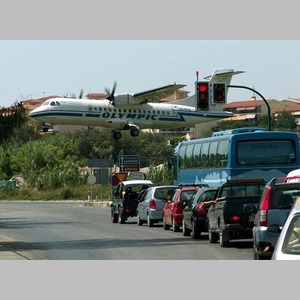} \\
        \parbox[][][c]{3.5in}{%
          GAN: a couch sitting in front of a house with a trash can \\
          CE: a white couch sitting in front of a house \\
          RL: a couch sitting in front of a house \\
          GT: a white couch on top of a grass curb with a black table in the background
        }
        &
        \parbox[][][c]{3.5in}{%
          GAN: a large passenger jet taking off from a busy street \\
          CE: a large passenger jet sitting on top of a runway \\
          RL: a group of cars parked on the runway at an airplane \\
          GT: an airplane descends very close to traffic stuck at a red light
        }
      \end{TAB}
    }
  \end{subfigure}
  \caption{Cherry-picked examples on the Out of Context (OOC) set.}
  \label{fig:goodooc}
\end{figure*}

\begin{figure*}[htb]
  \centering
  \begin{subfigure}[t]{.9\linewidth}
    \centering
    \resizebox{0.9\linewidth}{!}{               
      \begin{TAB}(c){|c|c|}{|c|c|c|c|}
        \includegraphics[width=4in, height=4in, keepaspectratio]{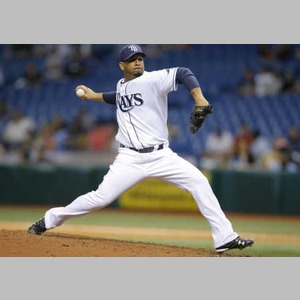} &
        \includegraphics[width=4in, height=4in, keepaspectratio]{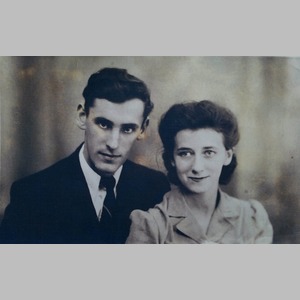} \\
        \parbox[][][c]{3.5in}{%
          GAN: a baseball player swinging a bat at a ball \\
          CE: a baseball player is swinging a bat at a ball \\
          RL: a baseball player swinging a bat at a ball \\
          GT: a man that is standing in the dirt with a glove
        }
        &
        \parbox[][][c]{3.5in}{%
          GAN: a black and white photo of two men in suits \\
          CE: a man and a woman standing next to each other \\
          RL: a black and white photo of a man and a woman \\
          GT: a man sitting next to a woman while wearing a suit
        }
        \\
        \includegraphics[width=4in, height=4in, keepaspectratio]{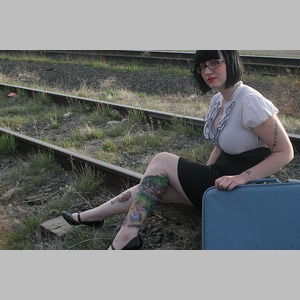} &
        \includegraphics[width=4in, height=4in, keepaspectratio]{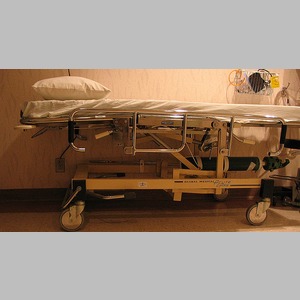} \\
        \parbox[][][c]{3.5in}{%
          GAN: a woman sitting on a bench looking at her cell phone \\
          CE: a woman sitting on a bench in a park \\
          RL: a woman sitting on the ground next to a bench \\
          GT: a woman is sitting with a suitcase on some train tracks
        }
        &
        \parbox[][][c]{3.5in}{%
          GAN: a bike that is in a room with a bike \\
          CE: a bicycle with a bicycle and a bicycle in a room \\
          RL: a room with a bed and a table in a room \\
          GT: the hospital bed is metal and has wheels
        }
      \end{TAB}
    }
  \end{subfigure}
  \caption{Lime-picked examples on the COCO test set.}
  \label{fig:badcoco}
\end{figure*}

\begin{figure*}[htb]
  \centering
  \begin{subfigure}[t]{.9\linewidth}
    \centering
    \resizebox{0.9\linewidth}{!}{               
      \begin{TAB}(c){|c|c|}{|c|c|c|c|}
        \includegraphics[width=4in, height=4in, keepaspectratio]{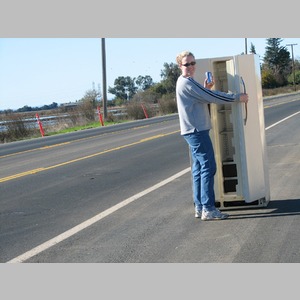} &
        \includegraphics[width=4in, height=4in, keepaspectratio]{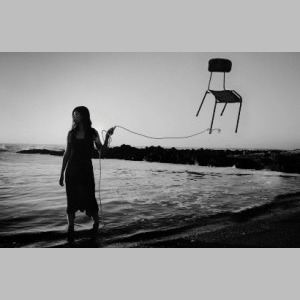} \\
        \parbox[][][c]{3.5in}{%
          GAN: a man standing on the side of the road with a skateboard \\
          CE: a man standing on the side of a road with a cell phone \\
          RL: a man standing on the side of a road with a cell phone \\
          GT: a woman holds a drink can while holding the door to a refrigerator that is sitting on the asphalt of a street
        }
        &
        \parbox[][][c]{3.5in}{%
          GAN: two people walking on a beach with a dog \\
          CE: a couple of people walking on a beach with a dog \\
          RL: a group of people walking on a beach with a dog \\
          GT: a lady is flying a chair as if its a kite while walking along the water edge
        }
        \\
        \includegraphics[width=4in, height=4in, keepaspectratio]{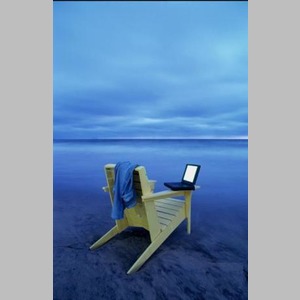} &
        \includegraphics[width=4in, height=4in, keepaspectratio]{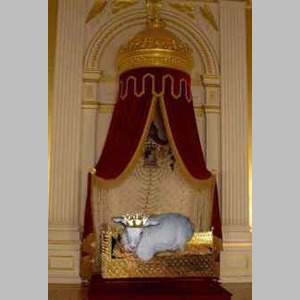} \\
        \parbox[][][c]{3.5in}{%
          GAN: a couple of chairs and a blue beach chairs on a beach \\
          CE: a couple of chairs sitting next to each other on a beach \\
          RL: a group of chairs and a table in the beach \\
          GT: a picture of a chair on an empty beach with a laptop on the arm
        }
        &
        \parbox[][][c]{3.5in}{%
          GAN: a painting of a vase in front of a fire hydrant \\
          CE: a painting of a fire place in a room \\
          RL: a bedroom with a bed and a clock on the wall \\
          GT: a white goat wearing a gold crown sits on a gold bed
        }
      \end{TAB}
    }
  \end{subfigure}
  \caption{Lime-picked examples on the Out of Context (OOC) set.}
  \label{fig:badooc}
\end{figure*}

In this section we present several examples of captions generated from our model. In particular, \Fig{fig:goodcoco} and \Fig{fig:goodooc} show captions for randomly picked images (from COCO and OOC respectively) which provide a good description of the image content.
We do the opposite in \Fig{fig:badcoco} and \Fig{fig:badooc} where examples of bad captions are provided for COCO and OOC respectively.

\end{document}